\documentclass[runningheads]{llncs}

\usepackage{eccv}

\usepackage{eccvabbrv}

\usepackage{graphicx}
\usepackage{booktabs}

\usepackage[accsupp]{axessibility}  

\usepackage{amsmath}
\usepackage{bm}
\usepackage{multirow} 
\usepackage{times}
\usepackage{epsfig}
\usepackage{graphicx}
\usepackage{amsmath}
\usepackage{amssymb}
\usepackage{booktabs}
\usepackage{multirow}
\usepackage{textcomp}
\usepackage{stfloats}
\usepackage{verbatim}
\usepackage{graphicx}
\usepackage{color,xcolor,colortbl}
\definecolor{lightgray}{gray}{0.95}
\definecolor{lightblue}{HTML}{95bddc}
\usepackage{bbding}
\usepackage{adjustbox}
\usepackage{makecell}
\usepackage{balance}

\usepackage[pagebackref,breaklinks,colorlinks,citecolor=eccvblue]{hyperref}

\usepackage{orcidlink}

\begin{document}

\title{VmambaIR: Visual State Space Model for Image Restoration} 





\author{Yuan Shi\inst{1,3}\thanks{Completed during internship at ByteDance.}  \and
Bin Xia\inst{2,3} \thanks{Yuan Shi and Bin Xia contributed equally to this paper.} \and
Xiaoyu Jin\inst{1} \and 
Xing Wang\inst{3} \and 
Tianyu Zhao\inst{3} \and 
Xin Xia\inst{3} \and 
Xuefeng Xiao\inst{3} \and 
Wenming Yang\inst{1}
}


\institute{Tsinghua Shenzhen International Graduate School, Tsinghua University \and
Department of Computer Science and Engineering, Chinese University of Hong Kong\and
Bytedance Inc.
\\\href{https://github.com/AlphacatPlus/VmambaIR}{VmambaIR Project}}

\maketitle

\vspace{-2mm}
\begin{abstract}
Image restoration is a critical task in low-level computer vision, aiming to restore high-quality images from degraded inputs. Various models, such as convolutional neural networks (CNNs), generative adversarial networks (GANs), transformers, and diffusion models (DMs), have been employed to address this problem with significant impact. However, CNNs have limitations in capturing long-range dependencies. DMs require large prior models and computationally intensive denoising steps. Transformers have powerful modeling capabilities but face challenges due to quadratic complexity with input image size. To address these challenges, we propose VmambaIR, which introduces State Space Models (SSMs) with linear complexity into comprehensive image restoration tasks. We utilize a Unet architecture to stack our proposed Omni Selective Scan (OSS) blocks, consisting of an OSS module and an Efficient Feed-Forward Network (EFFN). Our proposed omni selective scan mechanism overcomes the unidirectional modeling limitation of SSMs by efficiently modeling image information flows in all six directions. Furthermore, we conducted a comprehensive evaluation of our VmambaIR across multiple image restoration tasks, including image deraining, single image super-resolution, and real-world image super-resolution. Extensive experimental results demonstrate that our proposed VmambaIR achieves state-of-the-art (SOTA) performance with much fewer computational resources and parameters. Our research highlights the potential of state space models as promising alternatives to the transformer and CNN architectures in serving as foundational frameworks for next-generation low-level visual tasks.
  \keywords{State space models \and Mamba \and Image restoration }
\end{abstract}

\vspace{-2mm}
\section{Introduction}
\label{sec:intro}
Image restoration refers to the process of restoring high-quality images from their low-quality counterparts. This encompasses a range of tasks, including image deblurring, super-resolution, and image deraining. These tasks play a crucial role in the low-level domain of computer vision and have garnered significant attention over the years.

In recent years, deep learning techniques have made remarkable strides in the field of image restoration. 
Deep learning models, including convolutional neural networks (CNNs) \cite{dong2015image, dong2015compression, zhang2021plug, zhang2018image, xia2022efficient, yu2019free, zhang2017beyond}, generative adversarial networks (GANs) \cite{wang2018esrgan, gulrajani2017improved}, vision transformers \cite{chen2021pre, liang2021swinir, zamir2022restormer, chen2022simple}, and diffusion models (DMs) \cite{xia2023diffir, lin2023diffbir, wang2023exploiting}, have demonstrated exceptional capabilities in addressing complex image restoration tasks. These models leverage their ability to learn intricate patterns and features from extensive datasets, enabling them to effectively restore high-quality images from low-quality inputs.

However, CNNs often encounter limitations in modeling capabilities when dealing with large datasets and long-range dependencies. Diffusion models (DMs) in image restoration typically involve significant computational burden and time consumption due to the utilization of large prior models and intensive denoising processes. On the other hand, vision transformers exhibit a quadratic complexity in processing input sequences, which poses challenges in handling large-sized images which are commonly encountered in typical image restoration tasks.

Recently, state space model (SSM) \cite{gu2021combining, smith2022simplified, gu2023mamba}, a novel approach originating from control systems, has garnered attention due to its linear complexity in processing input sequences. However, state space models suffer from the limitations of unidirectional modeling of input data and a lack of spatial awareness. These issues pose challenges when it comes to handling visual data. The development of a comprehensive data modeling mechanism that effectively harnesses the multidimensional information embedded in visual data flow, coupled with the design of streamlined and efficient network architectures tailored for state space models, is a challenging and unresolved task. This endeavor aims to fully exploit the inherent linear complexity and powerful high-frequency modeling capabilities of state space models in image restoration tasks.
\begin{figure*}[tbp]
    \newlength\fsdurthree
    \setlength{\fsdurthree}{0mm}
    \Huge
    \centering
    \resizebox{1\linewidth}{!}{
            \begin{adjustbox}{valign=t}
                \begin{tabular}{ccc}
                    \includegraphics[width= \textwidth]{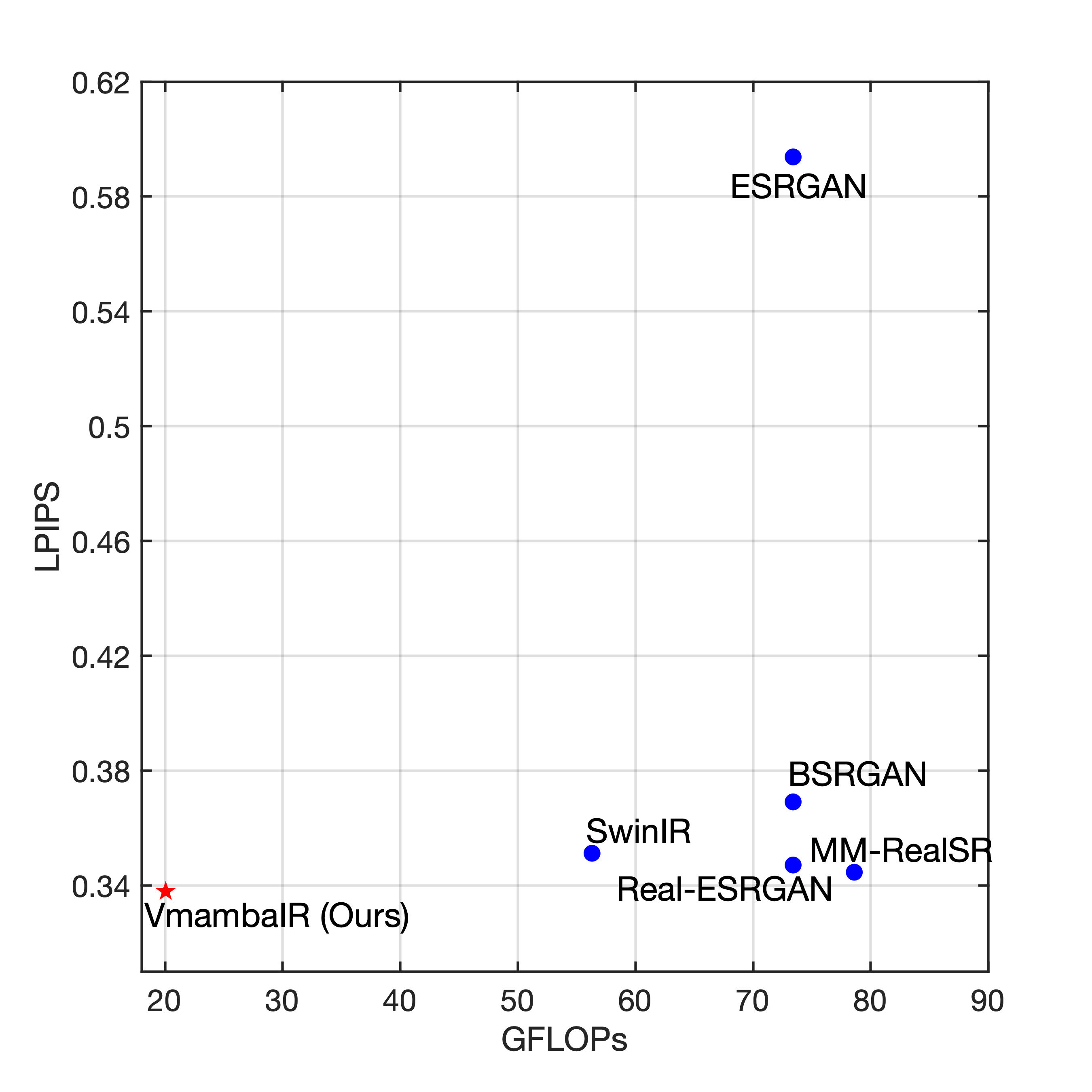} \hspace{\fsdurthree} &
                    \includegraphics[width= \textwidth]{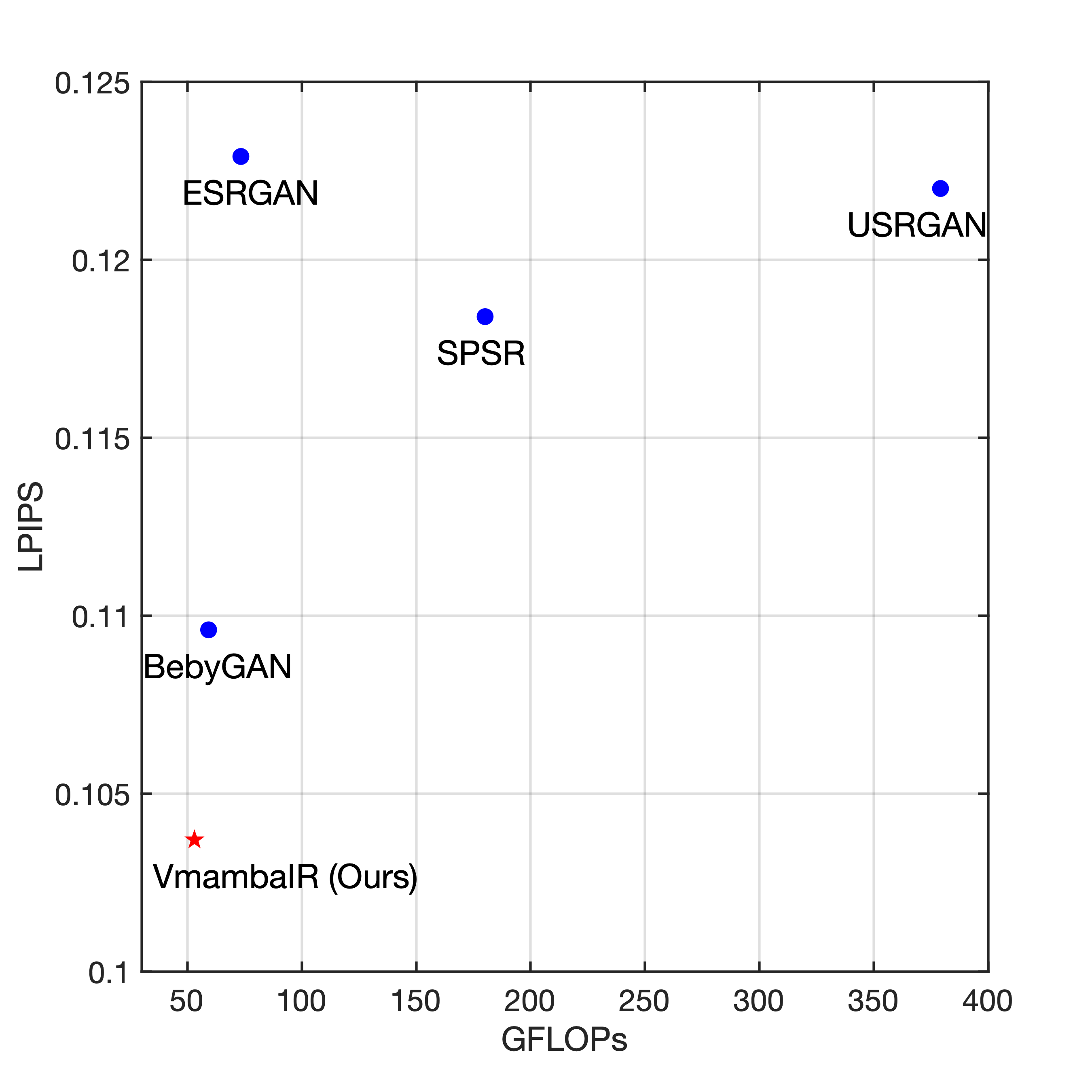} \hspace{\fsdurthree} &
                    \includegraphics[width= \textwidth]{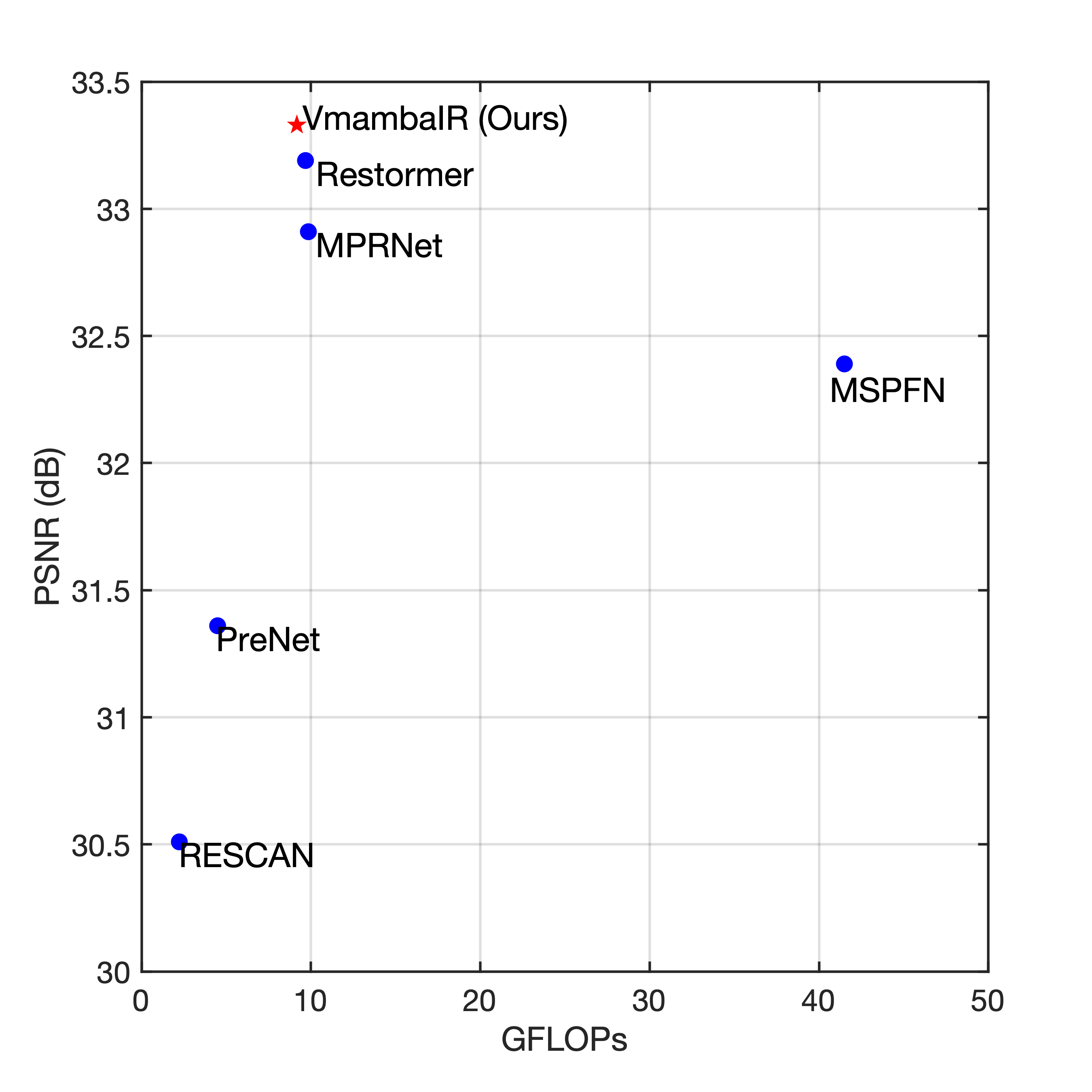} \hspace{\fsdurthree} 
                    \\
                    \makecell{(a) Real-world super-resolution} \hspace{\fsdurthree} &
                    \makecell{(b) Super-resolution} \hspace{\fsdurthree} &
                    \makecell{(c) Deraining}\hspace{\fsdurthree} 
                \end{tabular}
            \end{adjustbox}
    }
    \vspace{-1mm}
    \caption{Our VmambaIR demonstrates outstanding performance by achieving higher accuracy in image restoration tasks while requiring less computational cost. The GFLOPs are computed based on an input image size of $64\times64$. In the real-world super-resolution task, VmambaIR achieves higher reconstruction accuracy with only \textbf{26$\%$} of the computational cost.}
    \vspace{-3mm}
    \label{fig:com}
\end{figure*}

To tackle these challenges, we present VmambaIR, a comprehensive image restoration network that leverages the state space model. To exploit the multi-scale features of images more effectively, we developed a network architecture that draws inspiration from the Unet framework \cite{ronneberger2015u}. Our architecture incorporates a novel Omni Selective Scan (OSS) block, consisting of an OSS module and an Efficient Feed-Forward Network (EFFN). These components work together to enable a comprehensive and efficient modeling of the information flow. The omni selective scan, which serves as the core module of the OSS block, enables comprehensive modeling of the information flow, effectively addressing the limitation of unidirectional modeling in the mamba block \cite{gu2023mamba}.

We conducted extensive experiments on multiple image restoration tasks, including image deraining, single image super-resolution, and real-world image super-resolution. The experimental results, as shown in Fig. \ref{fig:com}, demonstrate that our proposed VmambaIR surpasses the accuracy of the current baseline on all image restoration tasks while requiring less computational resources. Particularly, in the real-world super-resolution task, VmambaIR achieves higher reconstruction accuracy with only \textbf{26$\%$} of the computational cost compared to the existing SOTA method.

Our contributions can be summarized as follows:
\begin{itemize}
\item[$\bullet$] We propose VmambaIR, a comprehensive image restoration model based on the state space model. VmambaIR incorporates our proposed OSS blocks into a Unet architecture, enabling effective handling of the multi-scale features of images.
\item[$\bullet$] Diverging from recent vision Mamba blocks \cite{ruan2024vm, liu2024vmamba, zheng2024u, zhu2024vision}, our designed OSS block comprises an OSS module and an EFFN. We have discovered that EFFN brings advantages to image restoration tasks, despite the fact that SSMs do not inherently require the handling of additional positional embedding bias.
\item[$\bullet$] We propose Omni Selective Scan, which enables comprehensive pattern recognition and modeling of the information flow in image data by modeling it from six directions, while incurring minimal additional computational burden.
\item[$\bullet$] We conducted experiments on comprehensive image restoration tasks, including single-image super-resolution, real-world image super-resolution, and image deraining. Extensive experimental results demonstrate that our VmambaIR achieves better performance with lower computational and parameter requirements.

\end{itemize}
 
\vspace{-2mm}
\section{Related work}
\vspace{-1mm}
\subsection{Image Restoration}
Image restoration, a long-standing and well-established research area in computer vision, has witnessed notable advancements with the emergence of deep learning. Throughout the years, a multitude of remarkable models and works have emerged, continually pushing the boundaries of image restoration techniques.

Pioneering works, such as SRCNN \cite{dong2015image} and ARCNN \cite{dong2015compression}, have employed compact convolutional neural networks (CNNs) to achieve impressive performance in image restoration tasks, particularly in the domain of image denoising. Subsequently, CNN-based methods gained popularity over traditional image restoration approaches. Over time, researchers have explored CNNs from various perspectives, leading to the development of more elaborate network architectures and learning schemes. These include the integration of residual blocks \cite{kim2016accurate, zhang2021plug}, the utilization of generative adversarial networks (GANs) \cite{wang2018esrgan, gulrajani2017improved}, the incorporation of attention mechanisms \cite{zhang2018image, xia2022efficient, dai2019second, yu2019free}, and the exploration of other innovative approaches \cite{jia2019focnet, fu2019jpeg, zeng2022aggregated, chen2022simple,xia2022knowledge,xia2022basic,jin2024llmra}.

Subsequently, the transformer, initially applied in natural language processing, has demonstrated exceptional performance in computer vision and image restoration domains. It has quickly become one of the fundamental model architectures for image restoration tasks \cite{chen2021pre, liang2021swinir, zamir2022restormer, chen2022simple}. The Transformer model outperforms CNNs by capturing global dependencies and modeling complex relationships. However, the computational complexity of self-attention in the transformer scales quadratically with the input image size. In image restoration tasks, where larger image sizes are often involved, this poses a challenge for the application of transformers.

Recently, the application of diffusion models in the field of image restoration has garnered attention\cite{xia2023diffir, lin2023diffbir, wang2023exploiting}. These models demonstrate powerful data-fitting capabilities, and pre-trained diffusion models possess extensive prior knowledge, enabling them to generate visually appealing restored images of high quality. However, the substantial computational resource requirements and limitations in fidelity have hindered the widespread adoption of diffusion models in the field of image restoration.
\vspace{-1mm}
\subsection{State Space Models (SSMs)}
SSMs have recently gained prominence in deep learning, particularly in the domain of state space transformation \cite{gu2021combining, smith2022simplified}. These models draw inspiration from continuous state space models in control systems and show promise in addressing long-range dependency issues, as demonstrated by LSSL \cite{gu2021combining}. To mitigate the computational complexity associated with LSSL, S4 \cite{gu2021efficiently} proposes parameter normalization using a diagonal structure, offering an alternative to CNNs and Transformers while specifically focusing on modeling long-range dependencies. Consequently, multiple structured state space models have emerged. S5 \cite{smith2022simplified} introduces the MIMO SSM and efficient parallel scan into the S4 layer, presenting the new S5 layer. And the Gated State Space layer \cite{mehta2022long} enhances expressivity by incorporating additional gating units into the existing S4 layer.

More recently, a data-dependent SSM layer and a generic language model backbone called Mamba have been proposed \cite{gu2023mamba}. Mamba surpasses Transformers in terms of performance on large-scale real data, showcasing its effectiveness across various sizes and demonstrating linear scalability in sequence length. The advantages of Mamba, particularly its computational efficiency for large-scale image processing, make it highly significant for research in the yet unexplored field of image restoration.

\vspace{-2mm}
\section{Preliminaries: State Space Models }
SSMs, which draw inspiration from continuous systems, are linear time-invariant systems that map the input stimulation $x(t) \in \mathbb{R}^{L}$ to the output response $y(t)\in \mathbb{R}^{L}$. Mathematically, SSMs can be formulated as linear ordinary differential equations (ODEs),
\begin{align}
  h^{'}(t) & = \bm{A}h(t) + \bm{B}x(t) \label{ode1} \; \\
  y(t) & = \bm{C}h(t) + \bm{D}x(t) \label{ode2}
\end{align}
where $h(t) \in \mathbb{R}^{N}$ is a hidden state, $\bm{A} \in \mathbb{R}^{N \times N}$, $\bm{B} \in \mathbb{R}^{N}$ and $\bm{C} \in \mathbb{R}^{N}$ are the parameters for a state size N, and $\bm{D} \in \mathbb{R}^{1}$ represents the skip connection.

Following that, discrete versions of SSMs were proposed, which transformed the original continuous-time nature and made them applicable in machine learning. The discretization process converts the ODE into a discrete function and aligns the model with the sample rate of the underlying signal present in the input data $x(t) \in \mathbb{R}^{L \times D}$ .

The ODE (Eq. \ref{ode1}) can be discretized using the zeroth-order hold (ZOH) rule, which incorporates a timescale parameter $\Delta$ to convert the continuous parameters $\bm{A}$, $\bm{B}$ into discrete parameters $\bm{\overline{A}}$, $\bm{\overline{B}}$, which can be defined as follows:
\begin{align}
  h^{'}_{t} & = \bm{\overline{A}}h_{t-1} + \bm{\overline{B}}x_{t} \; \\
  y_{t} & = \bm{C}h_{t} + \bm{D}x_{t} \; \\
  \bm{\overline{A}} &= e^{\Delta \bm{A}}        \; \\
  \bm{\overline{B}} &= (\Delta \bm{A})^{-1}(e^{\Delta \bm{A}}- I) \cdot \Delta \bm{B}
  \label{ode3}
\end{align}
where $\Delta \in \mathbb{R}^{D}$ and $\bm{B}, \bm{C} \in \mathbb{R}^{D \times N}$.

In contrast to previous linear time-invariant (LTI) SSMs, our proposed Omni Mamba leverages the selection mechanism introduced in Mamba\cite{gu2023mamba}. This empowers it to effectively capture the characteristics of long-sequence signals using a straightforward architecture, resulting in notable computational efficiency and accuracy.

\vspace{-2mm}
\section{Method}

In this section, we initially present the inter-block structures and overall network pipeline of our designed VmambaIR. Subsequently, we present the fundamental block of our model, the OSS block, which consists of an OSS module and an EFFN. This block effectively utilizes mamba's high-frequency modeling capability to capture information flow. Lastly, we introduce our proposed omni selective scan, which overcomes the limitations of unidirectional modeling in the state space model by efficiently modeling image information flow from all three dimensions.
\vspace{-1mm}
\subsection{Model Architecture}
\begin{figure}[tb]
  \centering
  \includegraphics[width=\textwidth]{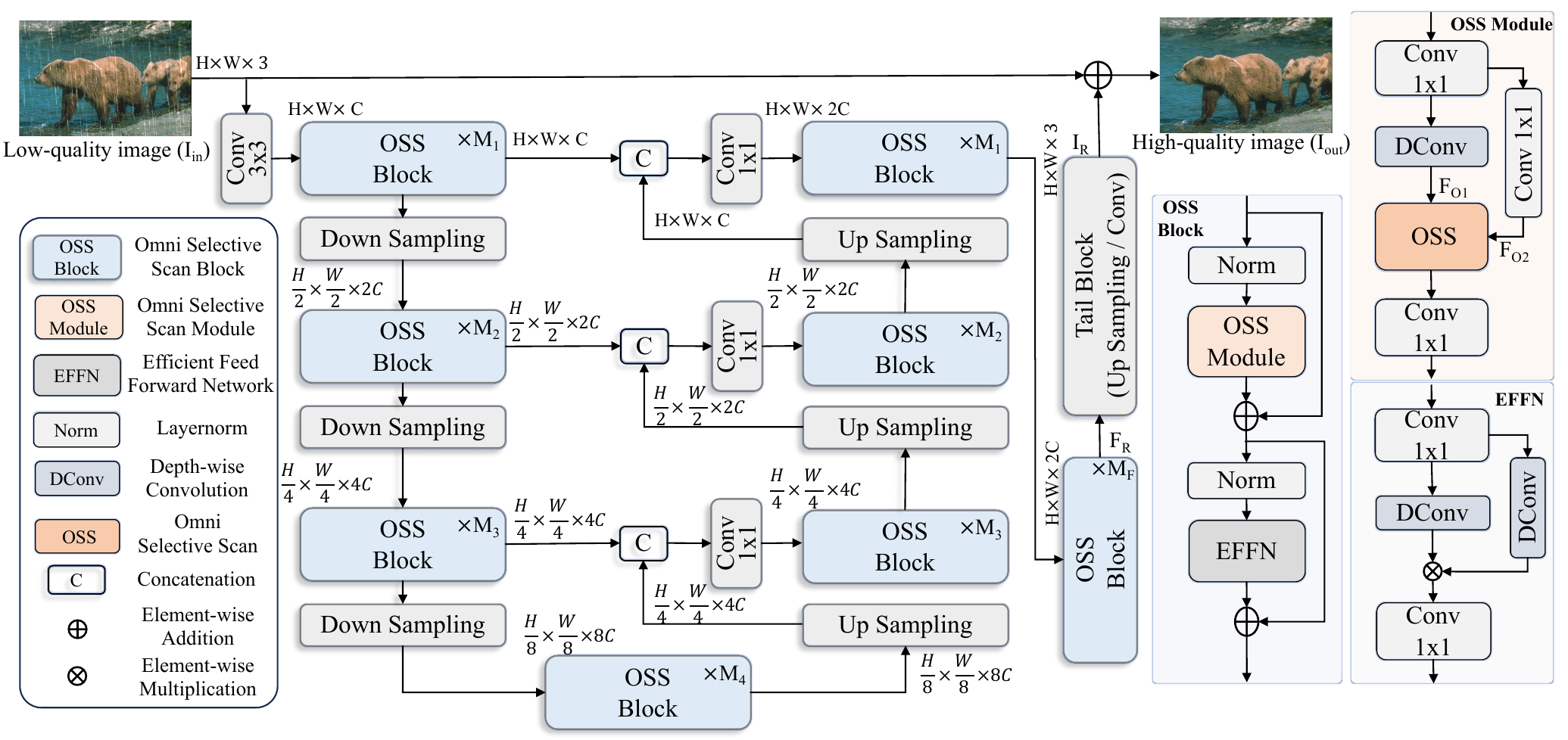}
  \vspace{-4mm}
  \caption{Overview of our VmambaIR. The low-quality image undergoes an initial convolutional processing step to extract shallow features. These features are then fed into a Unet architecture, which is constructed using our proposed OSS block, enabling the extraction and reconstruction of features at various scales. The reconstructed features are subsequently refined through multiple iterations of OSS blocks. Finally, the refined features are passed through a tail block, typically involving convolution or gradual upsampling, to reconstruct the final high-quality image.
  }
  \vspace{-3mm}
  \label{fig:model}
\end{figure}

Common inter-block structures in image restoration models include plain stacking structure\cite{liang2021swinir}, multi-stage architecture\cite{chen2021hinet}, multi-scale fusion architecture\cite{cho2021rethinking}, and UNet architecture\cite{zamir2022restormer, wang2022uformer}.
To ensure that VmambaIR is capable of capturing image features at different scales and remains robust to inputs of various sizes, we implemented a multi-scale UNet architecture based on our proposed OSS block, following \cite{ronneberger2015u, zamir2022restormer}. Specifically, the model architecture is depicted in Fig. \ref{fig:model}. 

Given a low-quality input image $I_{in} \in \mathbb{R}^{H \times W \times 3}$, a convolution is first applied to obtain the shallow feature embeddings $E_{1} \in \mathbb{R}^{H \times W \times C}$. Then the hierarchical encoding is achieved through three layers of OSS blocks and downsampling, applied to the features $E_{1}$.
The features $E_{2}$, $E_{3}$, and $E_{4}$, processed by each OSS block at different layers, are downsampled to sizes $\frac{H}{2} \times \frac{W}{2}$, $\frac{H}{4} \times \frac{W}{4}$, and $\frac{H}{8} \times \frac{W}{8}$ respectively. The features $E_{i}$ from the i-th layer of the encoder are concatenated with the features $D_{i+1}$ from the previous layer of the decoder through skip connections. After three upsampling and decoder layers, we obtain the output feature $D_{1}$ from the decoder. Subsequently, $D_{1}$ is refined through $M_{F}$ OSS blocks to obtain the feature $F_{R} \in \mathbb{R}^{H \times W \times 2C}$. 

For image restoration tasks such as image super-resolution that involve upsampling, we utilize convolution and pixel shuffle within the tail block to upsample the features $F_{R}$. On the other hand, for tasks like image de-raining or deblurring that do not require a change in image size, we directly process the feature $F_{R}$ using vanilla convolution to obtain the residual of the final output image $I_{R} \in \mathbb{R}^{H \times W \times 3}$. 
\vspace{-1mm}
\subsection{OSS Block}
The OSS block is employed to extract and model information flows from the omni domain of feature embeddings, as depicted in Figure \ref{fig:model}. In contrast to the recent design of the Vision Mamba block \cite{ruan2024vm, liu2024vmamba, zheng2024u, zhu2024vision}, which heavily relied on selection scanning mechanisms and linear mapping, our proposed OSS block introduces an OSS module capable of modeling information flows from diverse feature dimensions. Additionally, it incorporates an Efficient Feed-Forward Network (EFFN).
\vspace{-1mm}
\subsubsection{OSS Module}
The input of the OSS module is initially processed by a convolutional layer, generating two information flows. One flow undergoes refinement through depth-wise convolution and a Silu activation, capturing intricate patterns. Simultaneously, the other flow is processed with a Silu activation. The two flows enter the core OSS mechanism, which models information across all feature dimensions. Subsequently, the two flows are fused within the OSS, merging the refined features with complementary information. After passing through a 1x1 convolution, the output of OSS generates the final output of the OSS block, offering a comprehensive representation of the input with improved feature extraction and modeling capabilities.

It is noteworthy that, in the design of the OSS block, we opt for convolutional layers instead of linear layers to map the feature dimensions. This deliberate decision aims to minimize the operations conducted on feature shapes throughout the entire OSS block design. By leveraging convolutional layers, we not only enhance computational efficiency but also ensure a higher level of network-wide consistency.
\vspace{-1mm}
\subsubsection{Efficient Feed-Forward Network (EFFN)}
After the modeling process with the OSS module, the features are subjected to layer normalization to mitigate pattern collapse. Subsequently, the normalized features are passed through an efficient feed-forward network. Consistent with \cite{zamir2022restormer}, the EFFN structure, illustrated in Figure \ref{fig:model}, incorporates a 1x1 convolution to map the features into a high-dimensional space. The hidden layer features are then processed with depth-wise convolution and a gated mechanism. Lastly, a 1x1 convolution is employed to map the features back to their original dimension.

We have observed that the EFFN plays a crucial role in governing the information flow across the hierarchical levels in our pipeline, even though our omni selective scan does not necessitate positional encoding like self-attention \cite{liu2021swin, liang2021swinir} does. Through information flow regulation, the EFFN facilitates synchronized operation, enabling each level to contribute its specialized expertise and leverage the strengths of other levels. This ensures that each level focuses on capturing complementary fine details.
\vspace{-1mm}
\subsection{Omni Selective Scan Mechanism}
Mamba (S6) \cite{gu2023mamba} is an auto regressive model widely recognized for its effectiveness in temporal and causal sequence modeling. It excels at capturing sequence dependencies in a unidirectional manner with high efficiency. However, the causal processing of input data in Mamba limits its ability to capture information beyond the scanned portion. In contrast to Transformers, Mamba encounters difficulties in modeling non-causal relationships, such as those found in image data.

To address the unidirectional modeling limitation of Mamba, one straightforward approach is to process the input data simultaneously in both forward and backward directions \cite{zhu2024vision}. Alternatively, similar to vmamba \cite{liu2024vmamba}, images can be scanned in a two-dimensional plane. However, these methods fail to fully integrate channel dimension information, which is crucial for comprehensive image modeling and important in image processing.
UVM-Net \cite{zheng2024u} proposes a method that scans both the two-dimensional information and channel information of images. However, the scanning remains unidirectional, and the network requires a fixed input image size. This significantly limits the model's performance and convenience.

\begin{figure}[tb]
  \centering
  \includegraphics[width=\textwidth]{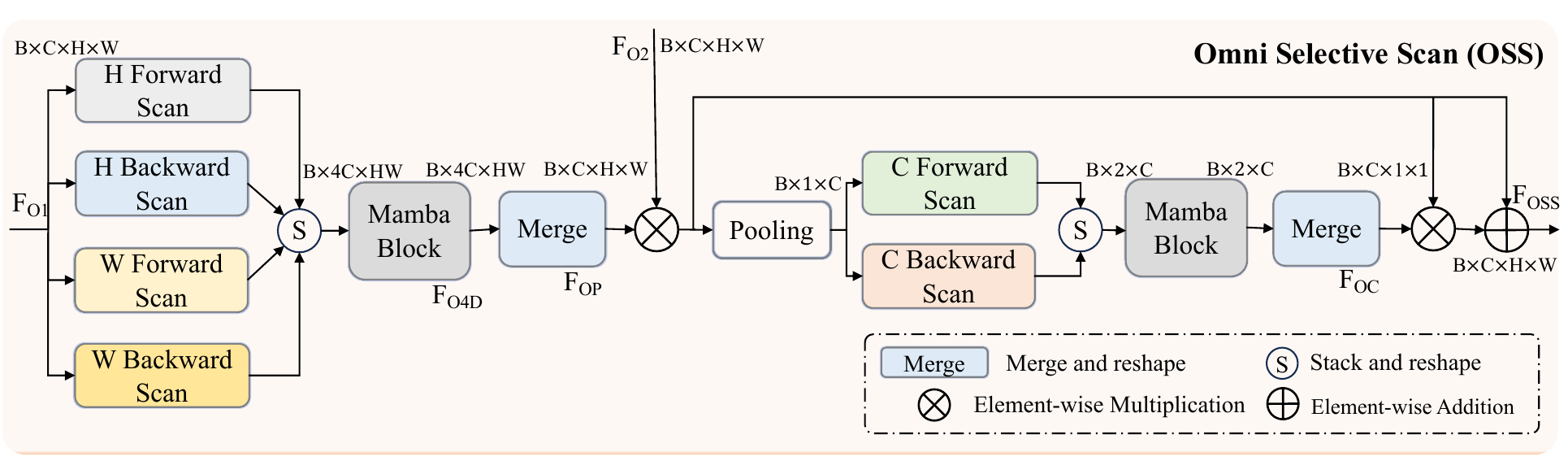}
  \vspace{-3mm}
  \caption{The architecture of one of our core designs, Omni Selective Scan. In the figure, "H Forward Scan", "W Forward Scan", and "C Forward Scan" indicate scanning from the top left to the bottom right, scanning from the bottom left to the top right on the two-dimensional image plane, and scanning the feature channels from front to back, respectively. The term "Backward" denotes the reverse direction of scanning. For the sake of simplicity in representation, operations on the feature dimensions, such as reshape and permute, have been omitted.
  }
  \vspace{-4mm}
  \label{fig:oss}
\end{figure}

To enhance the multidimensional modeling capability of Mamba in image processing, we introduce the Omni Selective Scan (OSS) mechanism, illustrated in Fig. \ref{fig:oss}.

We utilize the two information streams, $F_{O1}, F_{O2} \in \mathbb{R}^{B \times C \times H \times W } $ in OSS module, as inputs for the OSS. First, we perform bidirectional scanning in the longitudinal and transverse directions on $F_{O1}$ to capture the planar two-dimensional information of the features. "H Forward Scan" and "H Backward Scan" in Fig. \ref{fig:oss} refer to scanning from the top left to the bottom right and from the bottom right to the top left of the two-dimensional features, respectively. While "W forward scan" represents scanning the image from the bottom left to the top right. Afterward, we stack and reshape the features obtained from the four scans and model them using the Mamba block, resulting in $F_{O4D} \in \mathbb{R}^{B \times 4C \times HW}$. The features processed by the Mamba block are merged back into a single feature for planar modeling by splitting and adding the features from the four directions, resulting in the feature $F_{OP}\in \mathbb{R}^{B \times C \times H \times W } $. 

After multiplying $F_{OP}$ and $F_{O2}$, we first perform a pooling operation, and then scan the features in both forward and backward directions along the channel dimension. Compared to the channel modeling approach in \cite{zheng2024u}, our utilization of pooling operations not only significantly reduces computational complexity while maintaining nearly unchanged performance but also eliminates the requirement for fixed image input sizes in the model. Subsequently, we continue to employ bidirectional scanning to ensure comprehensive modeling of channel information. After undergoing sequential processing, including stacking, the Mamba block, and merging, we obtain the feature $F_{OC} \in \mathbb{R}^{B \times C \times 1 \times 1}$, which models the channel dimension. 

We employ a residual-based approach to effectively fuse the information from channel modeling and planar modeling, without the need for additional information streams. This strategy not only reduces computational costs but also enhances the stability of the model. Finally, we obtain the feature $F_{OSS} \in \mathbb{R}^{B \times C \times H \times W }$, which represents the comprehensive scanning and modeling of the input information flow by omni selective scan.

\begin{figure}[tb]
  \centering
  \includegraphics[width=0.7\textwidth]{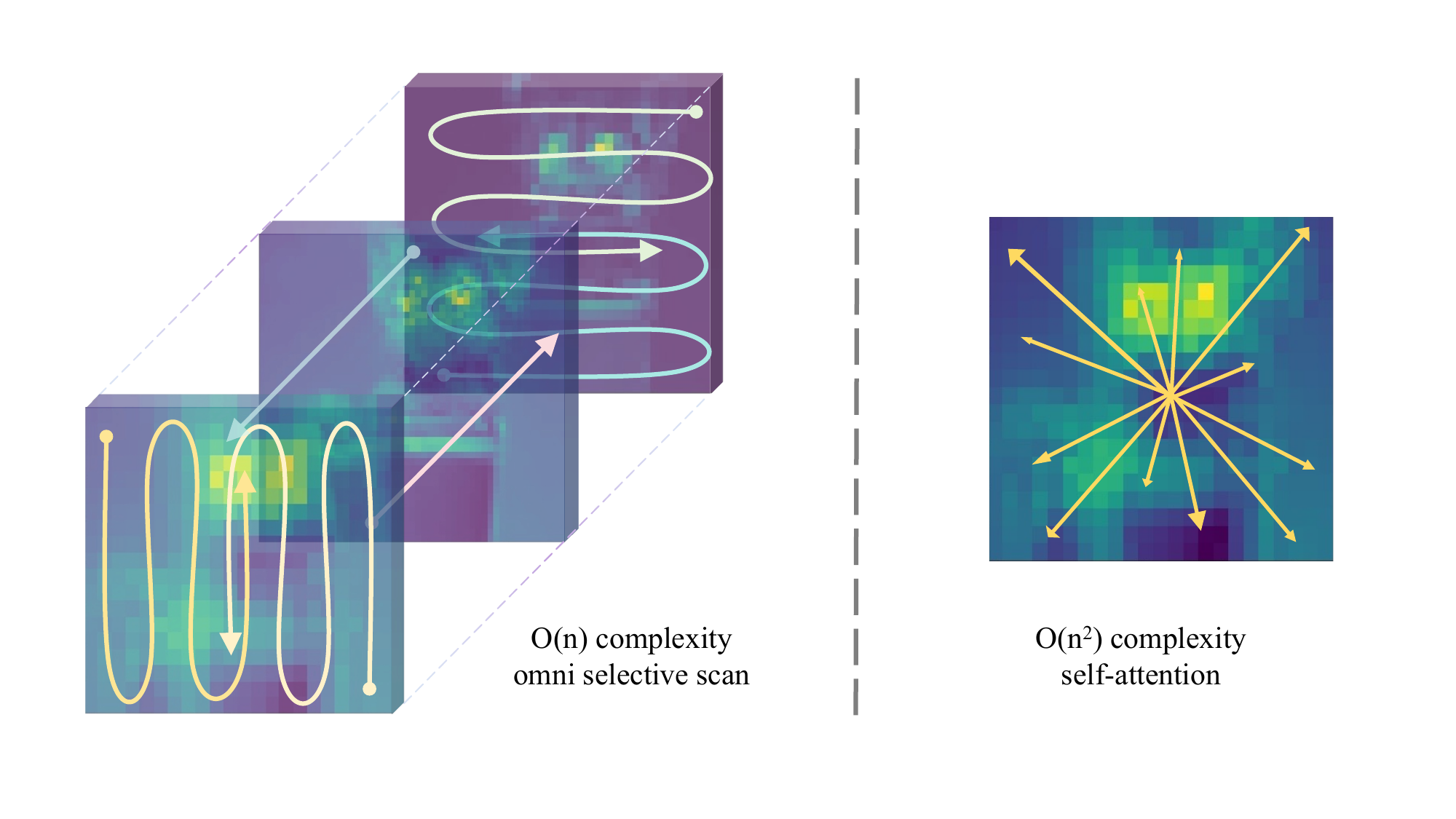}
  \vspace{-2mm}
  \caption{ The comparison between our proposed omni selective scan and self-attention reveals that Omni selective scan enables the modeling of image features from six directions and possesses linear computational complexity.
  }
  
  \label{fig:com2}
\end{figure}

Compared to the self-attention mechanism in Transformers and recent works related to vision mamba block, our proposed Omni Selective Scan (OSS) enables comprehensive modeling of image features from six directions with minimal additional computational burden. The process of scanning, modeling, and processing the information flow through our omni selective scan  can be illustrated in Fig. \ref{fig:com2}. 

While self-attention models operate within the two-dimensional plane and exhibit quadratic complexity with respect to the input sequence, our proposed omni selective scan leverages Mamba's long-range modeling capability to comprehensively capture three-dimensional image features while maintaining linear complexity.

\section{Experiments and Analysis}
\label{sec:blind}
\vspace{-1mm}
\subsection{Experiment Settings}
We validated the effectiveness of our proposed method on the following three image restoration tasks: (1) single image super-resolution, (2) real-world image super-resolution, and (3) image deraining.

We employed a four-layer symmetric encoder-decoder network structure for the aforementioned three tasks. The number of blocks in each layer varied as follows: [14, 1, 1, 1] for single image super-resolution, [6, 2, 2, 1] for real-world image super-resolution, and [4, 4, 6, 8] for deraining. Additionally, we integrated a refinement block with counts of 14, 6, and 2 for each respective task.  The feature dimensions of each layer in the network were set as [48, 96, 192, 384]. Following the training settings in \cite{zamir2022restormer}, we employed a progressive training strategy to handle input images of different sizes in the image deraining task. In image super-resolution tasks, the size of the ground-truth images is $256\times256$. We train models with Adam optimizer ($\beta_{1}$ = 0.9, $\beta_{2}$ = 0.99).

To achieve better visual reconstruction quality, we employed perceptual loss and GAN loss for image super-resolution tasks. For the other image restoration tasks, we utilized the L1 loss function as the training objective. 

\vspace{-1mm}
\subsection{Single Image Super-Resolution Results}
We trained our VmambaIR model on the DIV2K dataset (800 images) \cite{agustsson2017ntire} and Flickr2K dataset (2650 images) \cite{timofte2017ntire}. The batch sizes are set to 64, and the low-quality (LQ) patch size are $64\times64$ for $4\times$ super-resolution. We compared our method with existing GAN-based image super-resolution methods on various datasets, including Set14 \cite{zeyde2012single}, General100 \cite{dong2016accelerating}, Urban100 \cite{huang2015single}, Manga109 \cite{matsui2017sketch}, and DIV2K100 \cite{timofte2017ntire}. To comprehensively assess the visual quality and fidelity of the generated images, we evaluated the performance of our model using LPIPS and PSNR metrics. 
\begin{table*}[ht]
  \centering
  \caption{Quantitative comparison (LPIPS/PSNR) for 4$\times$ \textbf{Single image super-resolution} on benchmarks. Best and second best performance are marked in bold and underlined, respectively.}
  \vspace{-2mm}
   \resizebox{1\linewidth}{!}{
    \begin{tabular}{l|cccccccccc}
    \toprule[0.2em]
    \multirow{2}[2]{*}{\textbf{Method}} & \multicolumn{2}{c}{\textbf{Manga109}~\cite{matsui2017sketch}} & \multicolumn{2}{c}{\textbf{Set14}~\cite{zeyde2012single}} & \multicolumn{2}{c}{\textbf{General100}~\cite{dong2016accelerating}} & \multicolumn{2}{c}{\textbf{Urban100}~\cite{huang2015single}} & \multicolumn{2}{c}{\textbf{DIV2K100}~\cite{timofte2017ntire}} \\
          & LPIPS $\downarrow$& PSNR $\uparrow$& LPIPS $\downarrow$&PSNR $\uparrow$& LPIPS $\downarrow$&PSNR $\uparrow$& LPIPS $\downarrow$& PSNR $\uparrow$& LPIPS $\downarrow$&PSNR $\uparrow$\\
    \midrule
    SFTGAN~\cite{wang2018recovering} & 0.0716  & 28.17 & 0.1313 & 26.74 & 0.0947 & 29.16 & 0.1343 & 24.34 & 0.1331 & 28.09 \\
    SRGAN~\cite{ledig2017photo} & 0.0707 & 28.11 & 0.1327 & 26.84 & 0.0964 & 29.33 & 0.1439 & 24.41 & 0.1257 & 28.17 \\
    ESRGAN~\cite{wang2018esrgan} & 0.0649 & 28.41 & 0.1241 & 26.59 & 0.0879 & 29.43 & 0.1229 & 24.37 & 0.1154 & 28.18 \\
    USRGAN~\cite{zhang2020deep} & 0.0630 & 28.75 & 0.1347 & \underline{27.41} & 0.0937 & \underline{30.00} & 0.1330 & 24.89 & 0.1325 & \underline{28.79} \\
    SPSR~\cite{ma2020structure}  & 0.0672 & 28.56 & 0.1207 & 26.86 & 0.0862 & 29.42 & 0.1184 & 24.80 & 0.1099 & 28.18 \\
    BebyGAN~\cite{li2022best} & \underline{0.0529} & \underline{29.19} & \underline{0.1157} & 27.09 & \underline{0.0778} & 29.95 & \underline{0.1096} & \underline{25.23} & \underline{0.1022} & 28.62 \\
    \midrule[0.1em]
    VmambaIR (Ours) & \textbf{0.0496} & \textbf{29.99} & \textbf{0.1097} & \textbf{27.64} & \textbf{0.0762} & \textbf{30.34} & \textbf{0.1037} & \textbf{25.71} & \textbf{0.0995} & \textbf{29.18} \\
    
    \bottomrule[0.2em]
    \end{tabular}%
    }
  \label{tab:SR}%
  \vspace{-2mm}
\end{table*}%

\begin{figure*}[ht]
    \setlength{\fsdurthree}{0mm}
    \Huge
    \centering
   \resizebox{\linewidth}{!}{
        \begin{tabular}{cc}
            \begin{adjustbox}{valign=t}
                \Huge
                \begin{tabular}{c}
                    \includegraphics[height=1.9\textwidth]{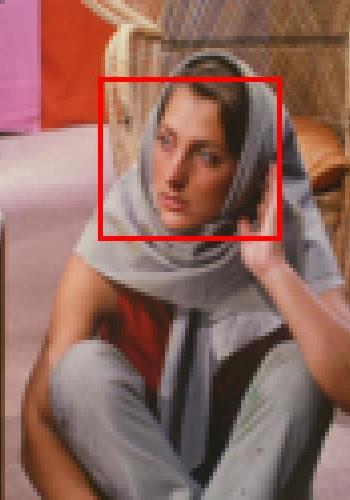} 
                \end{tabular}
                
            \end{adjustbox}
            
            \begin{adjustbox}{valign=t}
                \begin{tabular}{cccc}
                    \includegraphics[width= \textwidth]{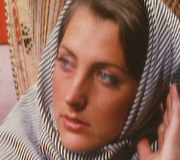} \hspace{\fsdurthree} &
                    \includegraphics[width= \textwidth]{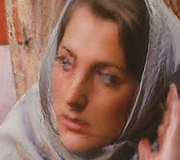} \hspace{\fsdurthree} &
                    \includegraphics[width= \textwidth]{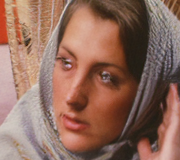} \hspace{\fsdurthree} 
                    \\
                    \Huge \scalebox{1.7}{HQ} \hspace{\fsdurthree} &
                    \makecell{\Huge \scalebox{1.7}{SRGAN}} \hspace{\fsdurthree} &
                    \makecell{\Huge \scalebox{1.7}{ESRGAN}} \hspace{\fsdurthree} 
                    \\
                    \includegraphics[width= \textwidth]{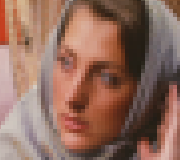} 
                    \hspace{\fsdurthree} &
                    \includegraphics[width= \textwidth]{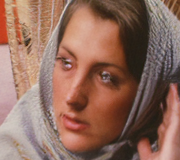} \hspace{\fsdurthree} &
                    \includegraphics[width= \textwidth]{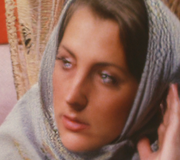} \hspace{\fsdurthree}  
                    \\ 
                    \Huge \scalebox{1.7}{LQ} \hspace{\fsdurthree} &
                    \Huge \scalebox{1.7}{BebyGAN} \hspace{\fsdurthree} &
                    \makecell{\Huge \scalebox{1.7}{VmambaIR (Ours)}} \hspace{\fsdurthree} 
                \end{tabular}
            \end{adjustbox}

            \begin{adjustbox}{valign=t}
                \Large
                \begin{tabular}{c}
                    \includegraphics[height=1.9\textwidth]{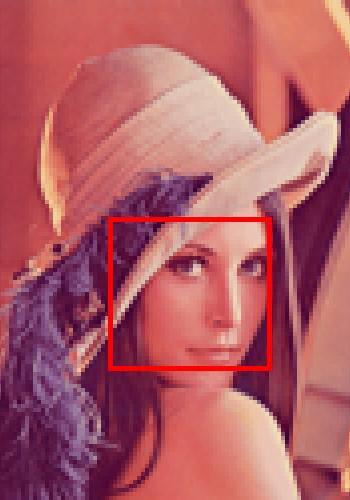} 
                \end{tabular}
                
            \end{adjustbox}
            
            \begin{adjustbox}{valign=t}
                \begin{tabular}{cccc}
                    \includegraphics[width= \textwidth]{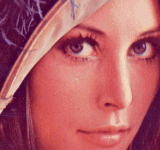} \hspace{\fsdurthree} &
                    \includegraphics[width= \textwidth]{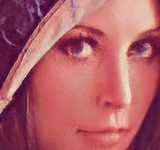} \hspace{\fsdurthree} &
                    \includegraphics[width= \textwidth]{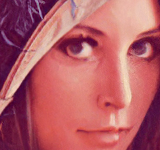} \hspace{\fsdurthree} 
                    \\
                    \Huge \scalebox{1.7}{HQ} \hspace{\fsdurthree} &
                    \makecell{\Huge \scalebox{1.7}{SRGAN}} \hspace{\fsdurthree} &
                    \makecell{\Huge \scalebox{1.7}{ESRGAN}} \hspace{\fsdurthree} 
                    \\
                    \includegraphics[width= \textwidth]{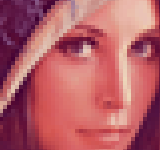} 
                    \hspace{\fsdurthree} &
                    \includegraphics[width= \textwidth]{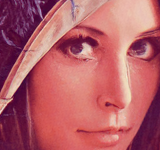} \hspace{\fsdurthree} &
                    \includegraphics[width= \textwidth]{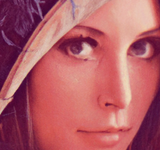} \hspace{\fsdurthree}  
                    \\ 
                    \Huge \scalebox{1.7}{LQ} \hspace{\fsdurthree} &
                    \Huge \scalebox{1.7}{BebyGAN} \hspace{\fsdurthree} &
                    \makecell{\Huge \scalebox{1.7}{VmambaIR (Ours)}} \hspace{\fsdurthree} 
                \end{tabular}
            \end{adjustbox}
            \\
            \begin{adjustbox}{valign=t}
                \Large
                \begin{tabular}{c}
                    \includegraphics[height=1.65\textwidth]{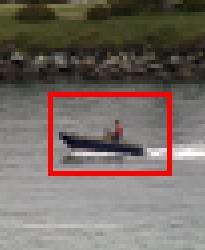} 
                \end{tabular}
                
            \end{adjustbox}
            
            \begin{adjustbox}{valign=t}
                \begin{tabular}{cccc}
                    \includegraphics[width= \textwidth]{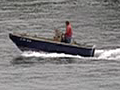} \hspace{\fsdurthree} &
                    \includegraphics[width= \textwidth]{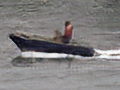} \hspace{\fsdurthree} &
                    \includegraphics[width= \textwidth]{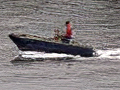} \hspace{\fsdurthree} 
                    \\
                    \Huge \scalebox{1.7}{HQ} \hspace{\fsdurthree} &
                    \makecell{\Huge \scalebox{1.7}{SRGAN}} \hspace{\fsdurthree} &
                    \makecell{\Huge \scalebox{1.7}{ESRGAN}} \hspace{\fsdurthree} 
                    \\
                    \includegraphics[width= \textwidth]{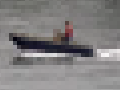} 
                    \hspace{\fsdurthree} &
                    \includegraphics[width= \textwidth]{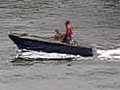} \hspace{\fsdurthree} &
                    \includegraphics[width= \textwidth]{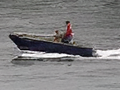} \hspace{\fsdurthree}  
                    \\ 
                    \Huge \scalebox{1.7}{LQ} \hspace{\fsdurthree} &
                    \Huge \scalebox{1.7}{BebyGAN} \hspace{\fsdurthree} &
                    \makecell{\Huge \scalebox{1.7}{VmambaIR (Ours)}} \hspace{\fsdurthree} 
                \end{tabular}
            \end{adjustbox}

            \begin{adjustbox}{valign=t}
                \Large
                \begin{tabular}{c}
                    \includegraphics[height=1.65\textwidth]{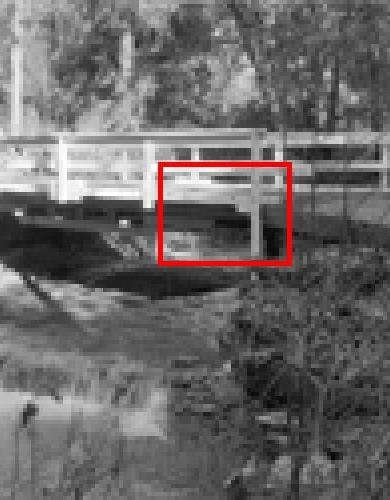} 
                \end{tabular}
                
            \end{adjustbox}
            
            \begin{adjustbox}{valign=t}
                \begin{tabular}{cccc}
                    \includegraphics[width= \textwidth]{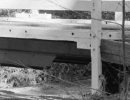} \hspace{\fsdurthree} &
                    \includegraphics[width= \textwidth]{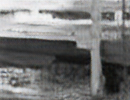} \hspace{\fsdurthree} &
                    \includegraphics[width= \textwidth]{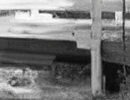} \hspace{\fsdurthree} 
                    \\
                    \Huge \scalebox{1.7}{HQ} \hspace{\fsdurthree} &
                    \makecell{\Huge \scalebox{1.7}{SRGAN}} \hspace{\fsdurthree} &
                    \makecell{\Huge \scalebox{1.7}{ESRGAN}} \hspace{\fsdurthree} 
                    \\
                    \includegraphics[width= \textwidth]{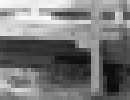} 
                    \hspace{\fsdurthree} &
                    \includegraphics[width= \textwidth]{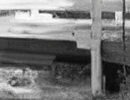} \hspace{\fsdurthree} &
                    \includegraphics[width= \textwidth]{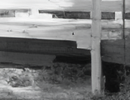} \hspace{\fsdurthree}  
                    \\ 
                    \Huge \scalebox{1.7}{LQ} \hspace{\fsdurthree} &
                    \Huge \scalebox{1.7}{BebyGAN} \hspace{\fsdurthree} &
                    \makecell{\Huge \scalebox{1.7}{VmambaIR (Ours)}} \hspace{\fsdurthree} 
                \end{tabular}
            \end{adjustbox}
        \end{tabular}
    }
    \caption{Visual comparison of \textbf{single image super-resolution} methods. Zoom-in for better details.}
    \label{fig:sisr}
    \vspace{-3mm}
\end{figure*}
The quantitative experimental results are presented in Table \ref{tab:SR}. It is evident that our VmambaIR outperforms existing SOTA methods in terms of both PSNR and LPIPS on all test datasets, demonstrating its superior performance in single-image super-resolution tasks. Specifically, on the Urban100 dataset, our method achieves a PSNR improvement of \textbf{0.48 dB} over BebyGAN \cite{li2022best} while maintaining a \textbf{lower} LPIPS score. To further validate the visual quality of the generated images by VmambaIR, the generated images from different methods are showcased in Fig. \ref{fig:sisr}.

From Figure \ref{fig:sisr}, it can be observed that the images generated by VmambaIR exhibit better fidelity and finer details, which aligns with the modeling capability of mamba in capturing high-frequency components in the information flow. In Figure \ref{fig:sisr}, our method is the only one that accurately generates the eyes and nose of the person without introducing any artifacts. Furthermore, the water surfaces, boats, and bridges in the images generated by our method also exhibit enhanced details and fewer artifacts.

\vspace{-1mm}
\subsection{Real-World Image Super-Resolution Results}
To further validate the image processing capabilities of our VmambaIR, we conducted experiments on more challenging real-world super-resolution tasks. We train our model on the DIV2K \cite{agustsson2017ntire}, Flickr2K \cite{timofte2017ntire} and OST v2 \cite{wang2018recovering} dataset for 4$\times$ real-world image super-resolution. we adopt the high-order degradation model following \cite{zhang2021designing, wang2021real} to generate degraded images. The low-quality (LQ) patch sizes are $64\times64$ in $4\times$ super-resolution.  
Similarly, we employ a GAN approach to train our model in this task, aiming to attain improved visual quality.
And we evaluate our VmambaIR on two benchmark datasets (NTIRE2020 \cite{lugmayr2020ntire} and AIM2019 \cite{lugmayr2019aim}) using LPIPS \cite{zhang2018unreasonable}, SSIM \cite{wang2004image} and PSNR. The quantitative results are shown in Table \ref{tab:realsr}.
\begin{table*}[t]
  \centering
  \caption{Quantitative comparison for  $4 \times$  \textbf{Real-World Image Super-Resolution} on benchmark datasets. Best and second best performance are marked in bold and underlined, respectively. The GFLOPs are computed based on an input image size of $64\times64$.}
  \vspace{-2mm}
  \resizebox{\linewidth}{!}{
    \begin{tabular}{l|cc|ccc|ccc}
    \toprule[0.2em]
    \multirow{2}[2]{*}{\textbf{Method}}&\multirow{1}[1]{*}{\textbf{Parms}}  &\multirow{1}[1]{*}{\textbf{FLOPs}} & \multicolumn{3}{c|}{\textbf{NTIRE2020}~\cite{lugmayr2020ntire}} & \multicolumn{3}{c}{\textbf{AIM2019}~\cite{lugmayr2019aim}} \\
    &(M) &(G) &\multicolumn{1}{c}{LPIPS $\downarrow$}  & \multicolumn{1}{c}{PSNR $\uparrow$} & \multicolumn{1}{c|}{SSIM $\uparrow$} & \multicolumn{1}{c}{LPIPS $\downarrow$}  & \multicolumn{1}{c}{PSNR $\uparrow$} & \multicolumn{1}{c}{SSIM $\uparrow$}  \\
    \midrule[0.2em]
    ESRGAN~\cite{zhang2021designing}&16.69 & 73.4 & \multicolumn{1}{c}{0.5938} & \multicolumn{1}{c}{21.14} &\multicolumn{1}{c|}{0.3119} & \multicolumn{1}{c}{0.5558}  & \multicolumn{1}{c}{23.17} &\multicolumn{1}{c}{0.6192}\\
    BSRGAN~\cite{zhang2021designing}&16.69 & 73.4 & \multicolumn{1}{c}{0.3691} & \multicolumn{1}{c}{\underline{26.75}} &\multicolumn{1}{c|}{0.7386} & \multicolumn{1}{c}{0.4048}  & \multicolumn{1}{c}{\textbf{24.20}} &\multicolumn{1}{c}{0.6904}\\
    Real-ESRGAN~\cite{wang2021real}&16.69 &73.4  & \multicolumn{1}{c}{0.3471} & \multicolumn{1}{c}{26.40} & \multicolumn{1}{c|}{0.7431} & \multicolumn{1}{c}{0.3956} & \multicolumn{1}{c}{23.89}&\multicolumn{1}{c}{0.6892} \\
    SwinIR~\cite{liang2021swinir}& 11.56 & 56.3  & \multicolumn{1}{c}{0.3512}  & \multicolumn{1}{c}{26.39}& \multicolumn{1}{c}{\underline{0.7414}}& \multicolumn{1}{c}{0.3980} & \multicolumn{1}{c}{23.88} & \multicolumn{1}{c}{0.6905}\\
    MM-RealSR~\cite{mou2022metric}& 26.13 & 78.6 & \multicolumn{1}{c}{\underline{0.3446}} & \multicolumn{1}{c}{25.19}&\multicolumn{1}{c|}{0.7404} & \multicolumn{1}{c}{0.3948}  & \multicolumn{1}{c}{23.05} &\multicolumn{1}{c}{0.6889}\\ 
    \midrule[0.1em]
    Vmamba-IR (Ours)& \textbf{10.50} & \textbf{20.5} & \multicolumn{1}{c}{\textbf{0.3379}}& \multicolumn{1}{c}{\textbf{27.06}}&  \multicolumn{1}{c|}{\textbf{0.7501}} & \multicolumn{1}{c}{\textbf{0.3891}} & \multicolumn{1}{c}{\underline{23.90}} &\multicolumn{1}{c}{\textbf{0.6972}} \\
    
    \bottomrule[0.2em]
    \end{tabular}%
    }
  \label{tab:realsr}%
  \vspace{-4mm}
\end{table*}%

\begin{figure*}[h]
    \setlength{\fsdurthree}{0mm}
    \LARGE
    \centering
   \resizebox{\linewidth}{!}{
        \begin{tabular}{c}
            \begin{adjustbox}{valign=t}
                \Large
                \begin{tabular}{c}
                    \includegraphics[height=1.75\textwidth]{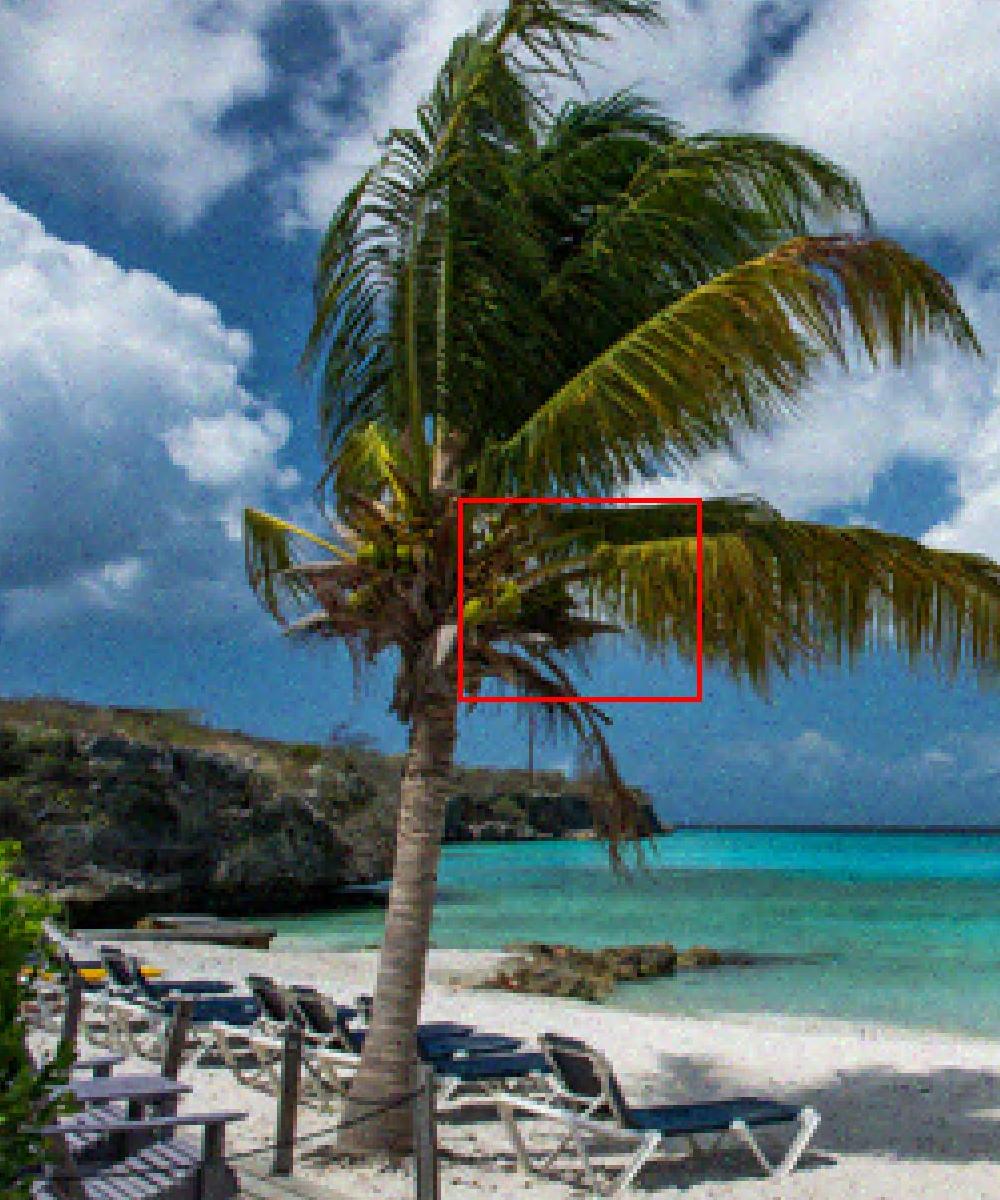} 
                \end{tabular}
                
            \end{adjustbox}
            
            \begin{adjustbox}{valign=t}
                \begin{tabular}{cccc}
                    \includegraphics[width= \textwidth]{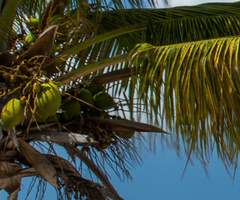} \hspace{\fsdurthree} &
                    \includegraphics[width= \textwidth]{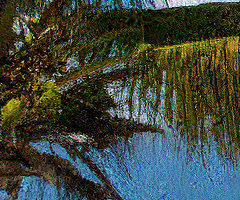} \hspace{\fsdurthree} &
                    \includegraphics[width= \textwidth]{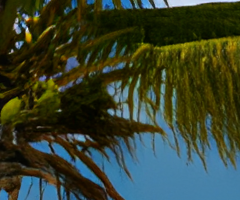} \hspace{\fsdurthree} &
                    \includegraphics[width= \textwidth]{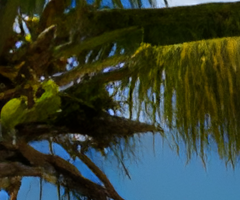} \hspace{\fsdurthree} 
                    \\
                    \Huge \scalebox{1.7}{HQ} \hspace{\fsdurthree} &
                    \Huge \scalebox{1.7}{ESRGAN} \hspace{\fsdurthree} &
                    \makecell{\Huge \scalebox{1.7}{SwinIR}} \hspace{\fsdurthree} &
                    \makecell{\Huge \scalebox{1.7}{Real-ESRGAN}} \hspace{\fsdurthree} 
                    \\
                    \includegraphics[width= \textwidth]{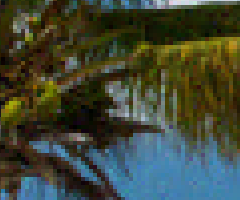} 
                    \hspace{\fsdurthree} &
                    \includegraphics[width= \textwidth]{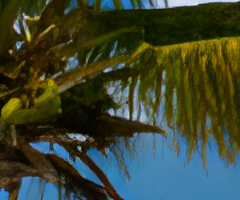} \hspace{\fsdurthree} &
                    \includegraphics[width= \textwidth]{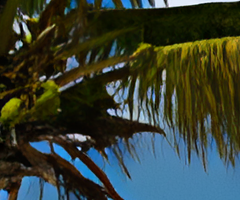} \hspace{\fsdurthree} &
                    \includegraphics[width= \textwidth]{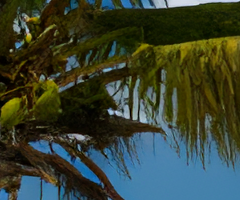} \hspace{\fsdurthree}  
                    \\ 
                    \Huge \scalebox{1.7}{LQ} \hspace{\fsdurthree} &
                    \Huge \scalebox{1.7}{BSRGAN} \hspace{\fsdurthree} &
                    \Huge \scalebox{1.7}{MM-RealSR} \hspace{\fsdurthree} &
                    \makecell{\Huge \scalebox{1.7}{VmambaIR (Ours)}} \hspace{\fsdurthree} 
                \end{tabular}
            \end{adjustbox}

            \\
            \begin{adjustbox}{valign=t}
                \Large
                \begin{tabular}{c}
                    \includegraphics[height=1.95\textwidth]{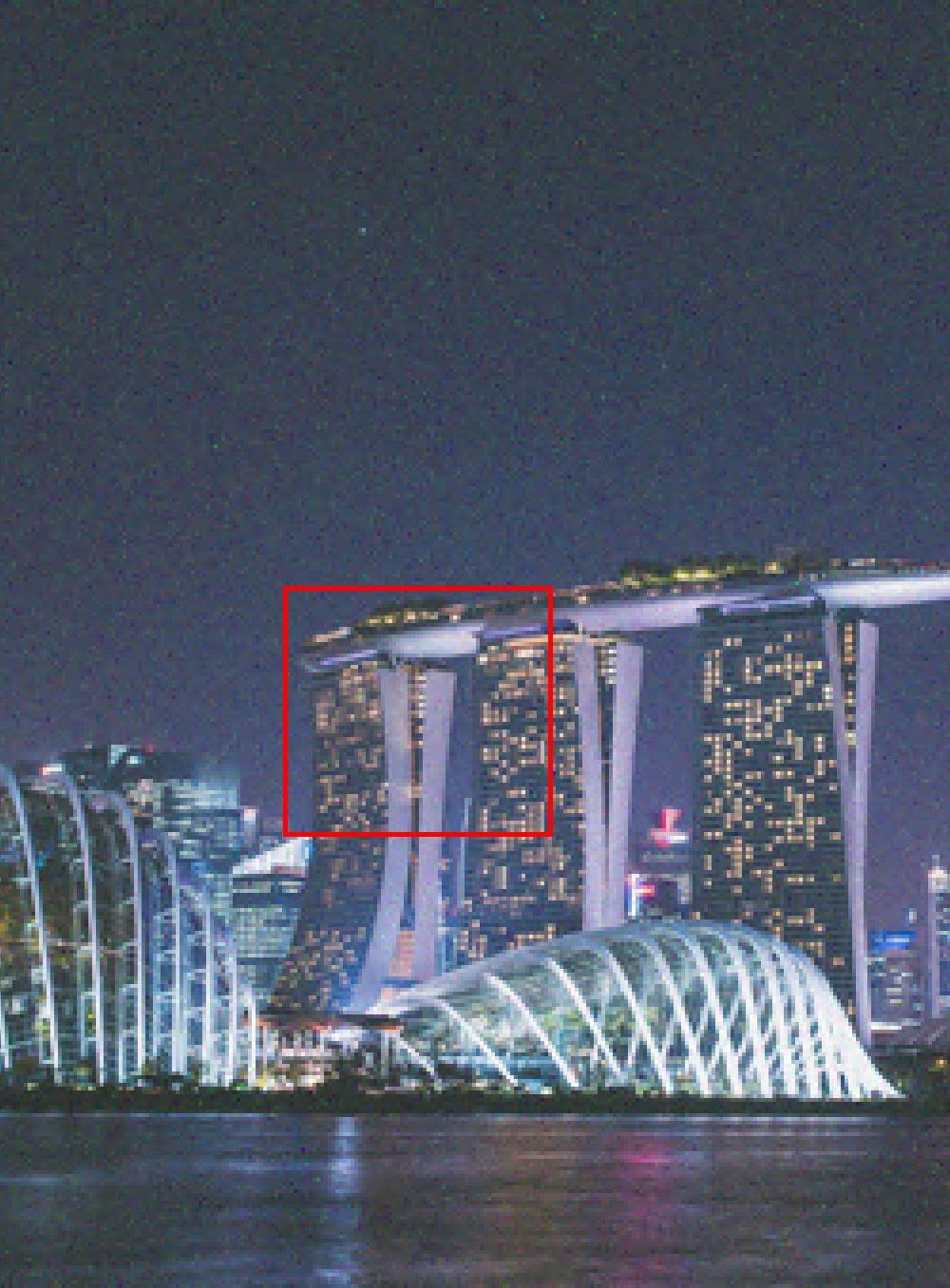} 
                \end{tabular}
                
            \end{adjustbox}
            
            \begin{adjustbox}{valign=t}
                \begin{tabular}{cccc}
                    \includegraphics[width= \textwidth]{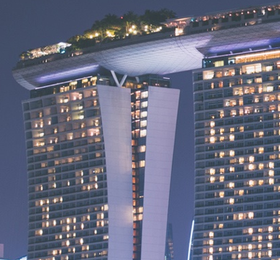} \hspace{\fsdurthree} &
                    \includegraphics[width= \textwidth]{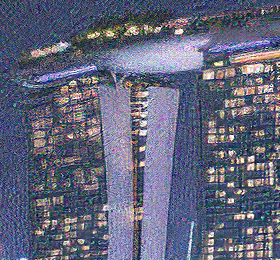} \hspace{\fsdurthree} &
                    \includegraphics[width= \textwidth]{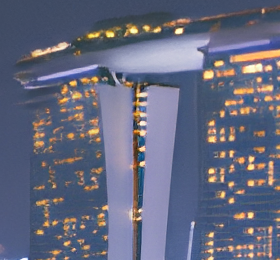} \hspace{\fsdurthree} &
                    \includegraphics[width= \textwidth]{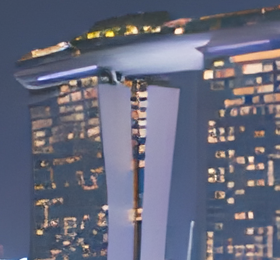} \hspace{\fsdurthree} 
                    \\
                    \Huge \scalebox{1.7}{HQ} \hspace{\fsdurthree} &
                    \Huge \scalebox{1.7}{ESRGAN} \hspace{\fsdurthree} &
                    \makecell{\Huge \scalebox{1.7}{SwinIR}} \hspace{\fsdurthree} &
                    \makecell{\Huge \scalebox{1.7}{Real-ESRGAN}} \hspace{\fsdurthree} 
                    \\
                    \includegraphics[width=\textwidth]{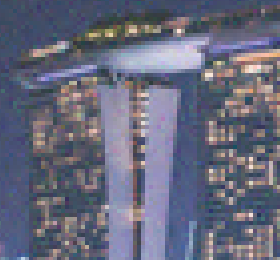} 
                    \hspace{\fsdurthree} &
                    \includegraphics[width= \textwidth]{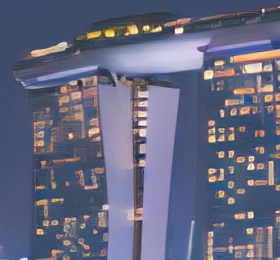} \hspace{\fsdurthree} &
                    \includegraphics[width= \textwidth]{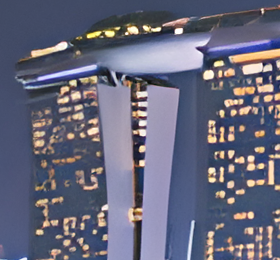} \hspace{\fsdurthree} &
                    \includegraphics[width= \textwidth]{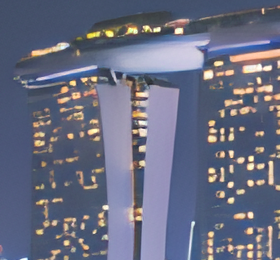} \hspace{\fsdurthree}  
                    \\ 
                    \Huge \scalebox{1.7}{LQ} \hspace{\fsdurthree} &
                    \Huge \scalebox{1.7}{BSRGAN} \hspace{\fsdurthree} &
                    \Huge \scalebox{1.7}{MM-RealSR} \hspace{\fsdurthree} &
                    \makecell{\Huge \scalebox{1.7}{VmambaIR (Ours)}} \hspace{\fsdurthree} 
                \end{tabular}
            \end{adjustbox}            
        \end{tabular}
    }
    \caption{Visual comparison of \textbf{real-world image super-resolution} methods. Zoom-in for better details.Our VmambaIR achieves superior high-frequency details and fidelity simultaneously.
}
    \label{fig:realsr}
    \vspace{-3mm}
\end{figure*}

The quantitative results demonstrate that our VmambaIR surpasses existing SOTA methods in comprehensive image quality evaluation metrics. Our VmambaIR consistently achieves lower LPIPS scores, and higher PSNR and SSIM scores across all test datasets, indicating its outstanding performance in real-world super-resolution tasks. 

Furthermore, our VmambaIR model showcases its robust modeling capacity in image restoration tasks by utilizing approximately \textbf{26}$\%$ of the computational resources and achieving the \textbf{lowest} parameter count compared to the existing state-of-the-art methods. Importantly, our approach does not rely on additional techniques such as distillation or pruning. Without a doubt, our VmambaIR showcases the immense potential of state space models in the realm of image restoration tasks.
To provide a more intuitive comparison, qualitative results of existing methods are presented in Fig. \ref{fig:realsr}.

The qualitative results of the real-world super-resolution task further highlight the powerful modeling capability of our VmambaIR in capturing high-frequency details in images. In the restored images by VmambaIR, the leaves and fruits of the coconut trees exhibit improved details. And our images are the only ones where both coconuts are recognized. Additionally, the high-rise building lights generated by VmambaIR exhibit fewer artifacts and do not suffer from excessive enhancement.

\vspace{-1mm}
\subsection{Image Deraining Results}
We trained and validate our VmambaIR on the Rain13K \cite{jiang2020multi}, Rain100H \cite{yang2017deep}, Rain100L \cite{yang2017deep}, Test1200 \cite{zhang2018density}, and Test2800 \cite{fu2017removing} dataset for image deraining task. Following the previous works \cite{purohit2021spatially, jiang2020multi,zamir2022restormer}, we evaluated the performance of our model in terms of PSNR and SSIM metrics on the Y channel in the YCbCr color space. And we employed a progressive learning strategy to train our model. The quantitative experimental results of our VmambaIR on image deraining are presented in Table \ref{tab:derain}. 

\begin{figure*}[htbp]
    \setlength{\fsdurthree}{0mm}
    \Huge
    \centering
   \resizebox{1\linewidth}{!}{
            \begin{adjustbox}{valign=t}
                \begin{tabular}{cccccc}

                    \includegraphics[width= \textwidth]{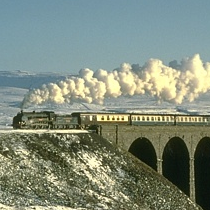} \hspace{\fsdurthree} &
                    \includegraphics[width=  \textwidth]{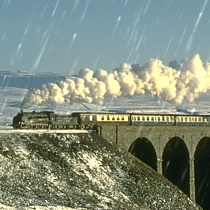} \hspace{\fsdurthree} &
                    \includegraphics[width=  \textwidth]{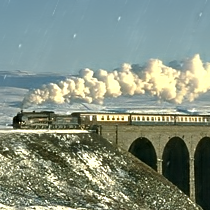} \hspace{\fsdurthree} &
                    \includegraphics[width=  \textwidth]{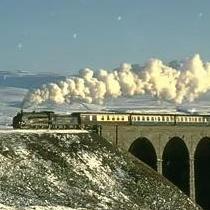} \hspace{\fsdurthree} &
                    \includegraphics[width= \textwidth]{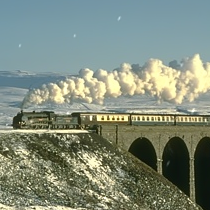} \hspace{\fsdurthree} &
                    \includegraphics[width=  \textwidth]{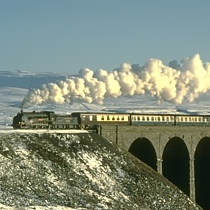} \hspace{\fsdurthree} 
                    \\
                    \includegraphics[width= \textwidth]{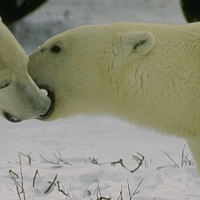} \hspace{\fsdurthree} &
                    \includegraphics[width=  \textwidth]{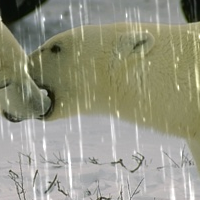} \hspace{\fsdurthree} &
                    \includegraphics[width=  \textwidth]{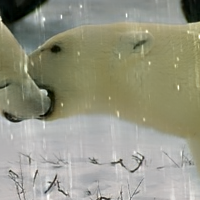} \hspace{\fsdurthree} &
                    \includegraphics[width=  \textwidth]{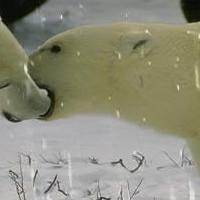} \hspace{\fsdurthree} &
                    \includegraphics[width= \textwidth]{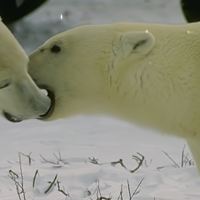} \hspace{\fsdurthree} &
                    \includegraphics[width=  \textwidth]{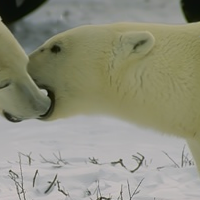} \hspace{\fsdurthree}  
                    \\
                    \scalebox{1.7}{HQ} \hspace{\fsdurthree} &
                    \scalebox{1.7}{LQ} \hspace{\fsdurthree} &
                    \scalebox{1.7}{\makecell{DerainNet~\cite{fu2017clearing}}} \hspace{\fsdurthree} &
                    \scalebox{1.7}{RESCAN~\cite{li2018recurrent}} \hspace{\fsdurthree} &
                    \scalebox{1.7}{Restormer~\cite{zamir2022restormer}} \hspace{\fsdurthree} &
                    \makecell{\scalebox{1.7}{VmambaIR (Ours)}} \hspace{\fsdurthree} 
                \end{tabular}
            \end{adjustbox}

    }
    \caption{Visual comparison of \textbf{image deraining} methods. Zoom-in for better details.}
    \vspace{-3mm}
    \label{fig:derain}
\end{figure*}

\begin{table*}[ht]
  \centering
  \caption{Quantitative comparison (PSNR/SSIM) for \textbf{Image Deraining} on five benchmark datasets. Best and second best performance are marked in bold and underlined, respectively.}
  \vspace{-2mm}
   \resizebox{1\linewidth}{!}{
    \begin{tabular}{l|cccccccc}
    \toprule[0.2em]
    \multirow{2}[2]{*}{\textbf{Method}}& \multicolumn{2}{c}{\textbf{Rain100H}~\cite{yang2017deep}} & \multicolumn{2}{c}{\textbf{Rain100L}~\cite{yang2017deep}} & \multicolumn{2}{c}{\textbf{Test2800}~\cite{fu2017removing}} & \multicolumn{2}{c}{\textbf{Test1200}~\cite{zhang2018density}} \\
         & PNSR $\uparrow$& SSIM $\uparrow$& PNSR $\uparrow$& SSIM $\uparrow$& PNSR $\uparrow$& SSIM $\uparrow$& PNSR $\uparrow$& SSIM $\uparrow$\\
    \midrule
    DerainNet~\cite{fu2017clearing}  & 14.92 & 0.592 & 27.03 & 0.884 & 24.31 & 0.861 & 23.38 & 0.835 \\
    UMRL~\cite{yasarla2019uncertainty}  &  26.01 & 0.832 & 29.18 & 0.923 & 29.97 & 0.905 & 30.55 & 0.910  \\
    RESCAN~\cite{li2018recurrent}  & 26.36 & 0.786 & 29.80 & 0.881 & 31.29 & 0.904 & 30.51 & 0.882\\
    PreNet~\cite{ren2019progressive}  & 26.77 & 0.858 & 32.44 & 0.950 & 31.75 & 0.916 & 31.36 & 0.911\\
    MSPFN~\cite{jiang2020multi}   & 28.66 &  0.860 &  32.40 & 0.933 & 32.82 & 0.930 & 32.39 & 0.916  \\
    MPRNet~\cite{zamir2021multi} & 30.41 & 0.890 & 36.40 & 0.965 & 33.64 & 0.938 & 32.91 & 0.916\\
    SPAIR~\cite{purohit2021spatially}   & 30.95 & 0.892 & 36.93 & 0.969 & 33.34 & 0.936 & 33.04 & 0.922\\
    Restormer~\cite{zamir2022restormer}   & \underline{31.46} &   \underline{0.904} & \underline{38.99} & \underline{0.978} & \textbf{34.18} & \underline{0.944} & \underline{33.19} & \underline{0.926} \\
    \midrule[0.2em]
    VmambaIR (Ours)  & \textbf{31.66} & \textbf{0.909} & \textbf{39.09} & \textbf{0.979} & \underline{34.01} & \textbf{0.944}  & \textbf{33.33} & \textbf{0.926} \\
    \bottomrule[0.2em]
    \end{tabular}%
    }
  \label{tab:derain}%
  \vspace{-2mm}
\end{table*}%
Clearly, our VmambaIR exhibits higher accuracy compared to existing SOTA methods, despite the inherent instability in image deraining tasks. In comparison to the existing method \cite{zamir2022restormer}, our VmambaIR achieves a PSNR improvement of \textbf{over} \textbf{0.1 dB} on the \textbf{Rain100H} \cite{yang2017deep}, \textbf{Rain100L} \cite{yang2017deep}, and \textbf{Test1200} \cite{zhang2018density} datasets, while maintaining less parameter and computational complexity. In addition, our method has consistently achieved higher SSIM scores on all datasets, indicating a higher visual quality of the restored images. To further validate the deraining performance of VmambaIR, we provide a qualitative comparison of the restored images from different methods in Fig. \ref{fig:derain}.

Compared to previous methods, our VmambaIR is capable of generating images in the image deraining task that are nearly indistinguishable from the ground truth. VmambaIR demonstrates almost perfect restoration results on the majority of the test images, highlighting the outstanding performance and generalization ability of our method.

\vspace{-1mm}
\subsection{Ablation Studies}
We conducted ablation experiments to investigate the effects of our proposed omni selective scan (OSS) mechanism, bidirectional channel scanning, efficient feed-forward network (EFFN), and OSS module detail improvements. We trained models with different network configurations using the L1 loss function on $4\times$ real-world image super-resolution tasks, and the results are presented in Table \ref{tab:ablation}.

\begin{table*}[ht]
  \centering
  \caption{The influence of different network configurations on our model. The PSNR results are evaluated on NTIRE2020 Track1 \cite{lugmayr2020ntire} for real-world image super-resolution. The performance and GFLOPs are measured on an LQ size of $64\times64$.}
  \vspace{-2mm}
  \resizebox{1\linewidth}{!}{
    \begin{tabular}{c|c|cccc|c}
    \toprule[0.2em]
    \multirow{1}[1]{*}{\textbf{Method}} & \multirow{1}[1]{*}{\textbf{\shortstack{GFLOPs}}} & \multirow{1}[1]{*}{\textbf{Plane scan}} & \multirow{1}[1]{*}{\textbf{Channel scan}} &  \multirow{1}[1]{*}{\textbf{EFFN}} &  \multirow{1}[1]{*}{\textbf{Conv1x1}}  & \multirow{1}[1]{*}{\textbf{PSNR}} \\
    \midrule
    VmambaIR (Ours) & 20.45  & \Checkmark    & \Checkmark     & \Checkmark    & \Checkmark     & 28.50 \\
    \midrule
    VmambaIR-V1 &   18.93         & \XSolidBrush     & \XSolidBrush     & \Checkmark   & \Checkmark &  28.07\\
    VmambaIR-V2 &    20.43    & \Checkmark   & \XSolidBrush     & \Checkmark     & \Checkmark   &  28.36\\
    VmambaIR-V3 &    20.03    & \Checkmark     & \Checkmark    & \XSolidBrush   & \Checkmark    &  28.39\\
    VmambaIR-V4 &   20.45      & \Checkmark      & \Checkmark     & \Checkmark    & \XSolidBrush     & 28.47 \\
    \bottomrule[0.2em]
    \end{tabular}%
}
  \label{tab:ablation}%
  \vspace{-2mm}
\end{table*}%
\vspace{-1mm}
\subsubsection{Effects of omni selective scan}
In VmambaIR-V1, we assessed the influence of the proposed Omni Selective Scan on the network. During training, we employed one-way scanning in the image features instead of our omni selective scan. As shown in Table \ref{tab:ablation}, the network's computational complexity reduced by approximately 7$\%$, while the network's accuracy experienced a decline of approximately \textbf{0.43} dB. This observation underscores the substantial limitations imposed by the one-way modeling of the state space model in processing two-dimensional image information.
\vspace{-1mm}
\subsubsection{Effects of bidirectional channel scanning}
In VmambaIR-V2, we employed 2D selective scan instead of Omni Selective Scan to validate the impact of the channel scanning mechanism in Omni Selective Scan. By removing the bidirectional channel scanning, the computational complexity of the network remained almost the same, but the network's accuracy decreased by approximately \textbf{0.14} dB, which demonstrates that the channel scanning we designed is very efficient and effective.
\vspace{-1mm}
\subsubsection{Effects of efficient feed-forward network}
In VmambaIR-V3, we abandoned the EFFN module in the OSS block and solely utilized the OSS module for network modeling. To maintain comparable network scale, we increased the number of OSS blocks. With similar computational complexity, VmambaIR with EFFN achieved higher accuracy, demonstrating the effectiveness of our OSS block and EFFN. Additionally, due to significant data type and dimension conversions in the current selective scan operations, the speed of selective scan operations with the same computational complexity is slower than vanilla convolution. Therefore, we employed EFFN to enhance both the computational accuracy and efficiency of our proposed VmambaIR.
\vspace{-1mm}
\subsubsection{Improvements in OSS module}
In VmambaIR-V4, we conducted a validation study on the influence of specific detailed design choices within the OSS module on the network. In VmambaIR-V4, we opted to use linear layers and additional reshape operations for feature dimension transformations, as opposed to 1x1 convolutions. Despite maintaining comparable computational complexity, parameter count, and accuracy, our designed VmambaIR demonstrated an approximate \textbf{8.6}$\%$ enhancement in computational speed.
\vspace{-1mm}
\section{Conclusion}
In this paper, we propose VmambaIR, a novel image restoration network, leveraging the linear complexity and high-frequency modeling capabilities of the mamba block \cite{gu2023mamba}. We employ the UNet \cite{ronneberger2015u} architecture to enable our proposed OSS block to effectively model and utilize images at different scales. The OSS block consists of an OSS module and an EFFN module. The OSS module leverages mamba's long-range modeling capability to comprehensively and efficiently model image features. The EFFN further maps and modulates the image information flow, enhancing the accuracy and efficiency of the network. Our model's capabilities are evaluated on various image restoration tasks, including image deraining, image super-resolution, and real-world image super-resolution. Extensive experimental results demonstrate that our designed model achieves state-of-the-art performance. Furthermore, our network is intentionally designed without the utilization of elaborate techniques such as distillation, teacher networks, hybrid network structures, and others. Our primary objective is to establish a simple yet effective mamba image restoration baseline that can serve as both inspiration and motivation for future research endeavors. Through the demonstration of the potential of state space models in the field of image restoration, our work endeavors to make a substantial and valuable contribution to the progress and development of this domain.

%
%
\bibliographystyle{splncs04}
\bibliography{main}

\newpage
\section{Appendix}

\subsection{More Training Details on Real-World Image Super-Resolution}
To generate data and train the network, we adopted a configuration identical to that of Real-ESRGAN~\cite{wang2021real}. Specifically, we implemented a dual degradation process to produce low-quality images. This involved applying random resizing, Gaussian noise, gray noise, and blur techniques to introduce various forms of degradation. Throughout this process, we maintained consistency with the degradation parameter settings used in Real-ESRGAN~\cite{wang2021real} to ensure comparable results.

The ground-truth images were cropped to a size of $256 \times 256$. And the size of training images were kept same in the training process for image super-resolution tasks. During the training process, we utilized 8 V100 GPUs, with each GPU processing a batch size of 9. To optimize the network, we employed the Adam optimizer with specific parameter values, namely $\beta_{1}$ = 0.9 and $\beta_{2}$ = 0.99. The initial learning rate was set to $1\times e^{-4}$, and a weight decay of 0 was applied. To control the learning rate, we employed a MultiStepLR strategy with a decay factor $\gamma$ of 0.5. The model was trained for 300,000 iterations. On average, the training process for the $4 \times$ real-world GAN-based image super-resolution model took approximately 5 days.

For training our VmambaIR, we incorporated GAN loss, perceptual loss, and L1 loss. These losses were given equal weights during the entire training process. To serve as the discriminator in our network, we adopted a U-Net architecture with a feature dimension of 64. The optimizer settings used for the generator were also applied to the Unet discriminator, ensuring consistency across both components.

\subsection{More Training Details on Single Image Super-Resolution}
For the GAN-based single image super-resolution task, we generated low-quality training images by employing bicubic downsampling. Following prior works, we utilized the DF2K dataset, which consists of 3450 images, to train our model.

For single image super-resolution task, the ground-truth images were cropped to a size of $256 \times 256$. and the low-quality images were $64 \times 64$ for $4 \times$ super-resolution. During the training process, we utilized 8 V100 GPUs, with each GPU processing a batch size of 8. To optimize the network, we employed the Adam optimizer with specific parameter values, namely $\beta_{1}$ = 0.9 and $\beta_{2}$ = 0.99. The initial learning rate was set to $2\times e^{-4}$, and a weight decay of 0 was applied. To control the learning rate, we employed a MultiStepLR strategy with a decay factor $\gamma$ of 0.5. The model was trained for 300,000 iterations. On average, the training process for the $4 \times$ GAN-based image super-resolution model took approximately 5 days.

Similar to the settings in real-world image super-resolution, during the entirety of the training process, we incorporated GAN loss, perceptual loss, and L1 loss into our model. To ensure equitable impact from each loss function, we assigned them equal weights. And the learning rate for the discriminator Unet was configured to $1\times e^{-4}$.

\subsection{More Training Details on Image Deraining}
Unlike in image super-resolution tasks, in order to capture deeper features of images for the image de-raining task, the number of blocks per layer in the network was set to [4, 4, 6, 8], and the number of refinement blocks was set to 2.

In accordance with the progressive training approach outlined in Restormer \cite{zamir2022restormer}, we followed specific settings during the training process. The input image sizes were sequentially set as [128, 160, 192, 256, 320, 384], while the corresponding batch sizes per GPU were [8, 5, 3, 2, 1, 1]. To train VmambaIR for the image deraining task, we utilized a total of 8 V100 GPUs. The training procedure spanned approximately 6 days, encompassing the complete training of our model.

We employed L1 loss as the objective function for training the image deraining task. The optimization was performed using the AdamW optimizer with  $\beta_{1}$ = 0.9 and $\beta_{2}$ = 0.999. The initial learning rate for training was set to $3\times e^{-4}$, and a weight decay of $1\times e^{-4}$ was applied. To control the learning rate during training, we utilized the cosine annealing technique, following the settings in restormer \cite{zamir2022restormer}.

\subsection{More Visual Comparisons on Real-World Image Super-Resolution}
\begin{figure*}[h]
    \setlength{\fsdurthree}{0mm}
    \LARGE
    \centering
   \resizebox{0.95\linewidth}{!}{
        \begin{tabular}{c}
            \begin{adjustbox}{valign=t}
                \Large
                \begin{tabular}{c}
                    \includegraphics[height=2.15\textwidth]{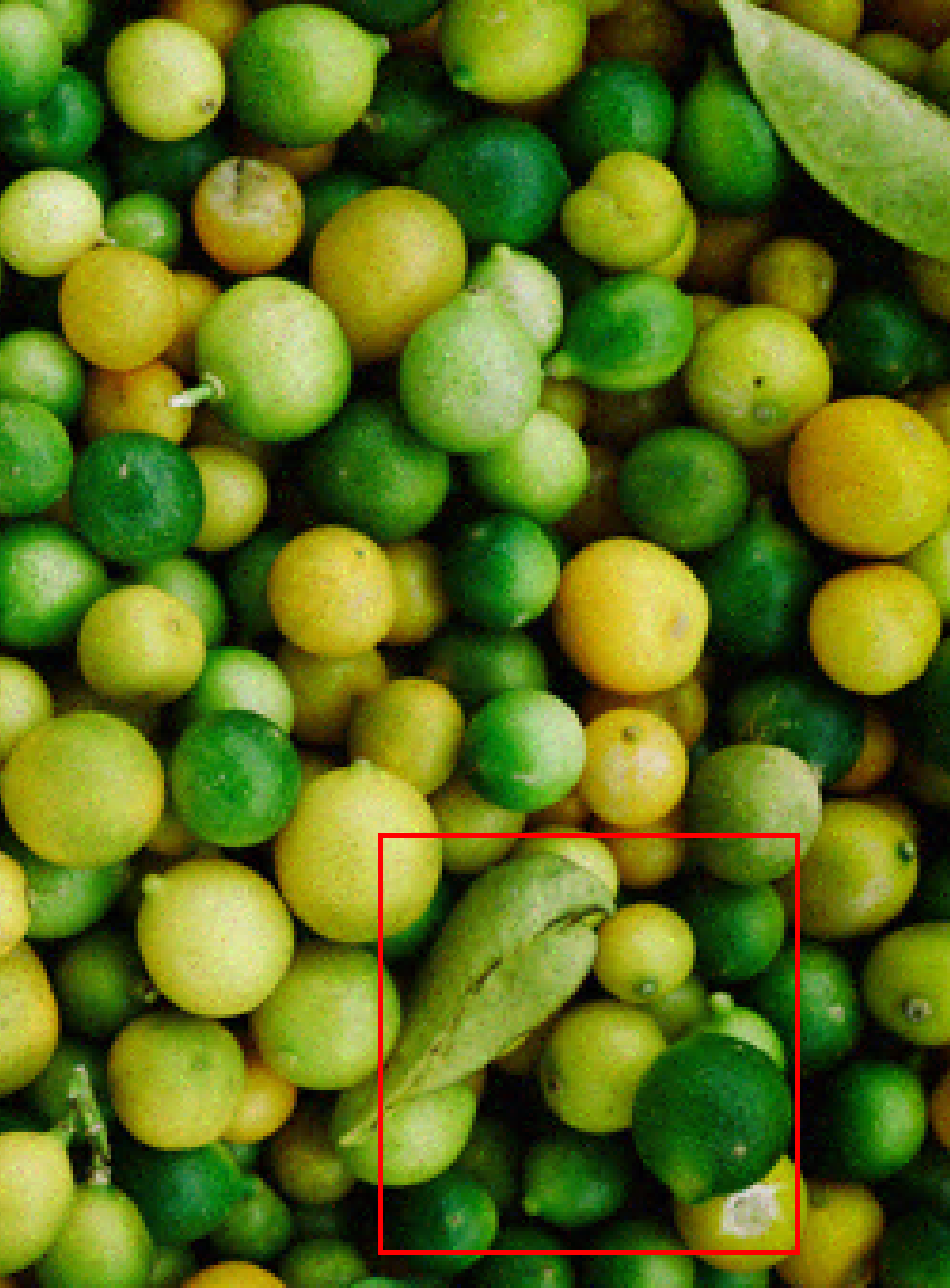} 
                \end{tabular}
                
            \end{adjustbox}
            
            \begin{adjustbox}{valign=t}
                \begin{tabular}{cccc}
                    \includegraphics[width= \textwidth]{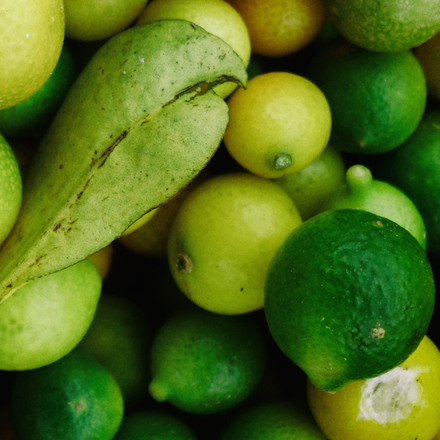} \hspace{\fsdurthree} &
                    \includegraphics[width= \textwidth]{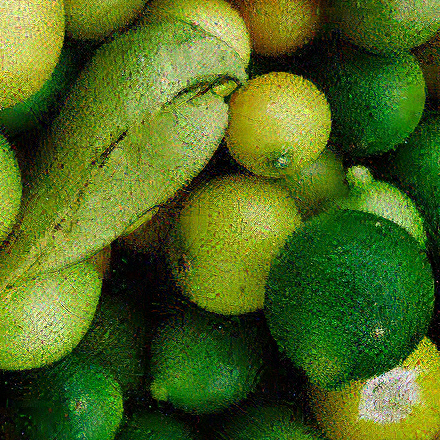} \hspace{\fsdurthree} &
                    \includegraphics[width= \textwidth]{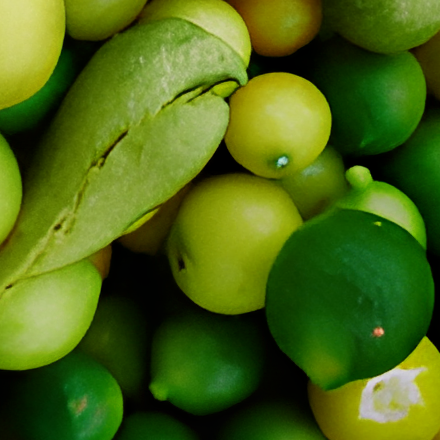} \hspace{\fsdurthree} &
                    \includegraphics[width= \textwidth]{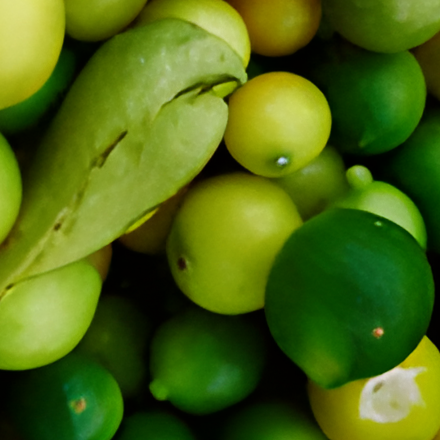} \hspace{\fsdurthree} 
                    \\
                    \Huge \scalebox{1.7}{HQ} \hspace{\fsdurthree} &
                    \Huge \scalebox{1.7}{ESRGAN} \hspace{\fsdurthree} &
                    \makecell{\Huge \scalebox{1.7}{SwinIR}} \hspace{\fsdurthree} &
                    \makecell{\Huge \scalebox{1.7}{Real-ESRGAN}} \hspace{\fsdurthree} 
                    \\
                    \includegraphics[width= \textwidth]{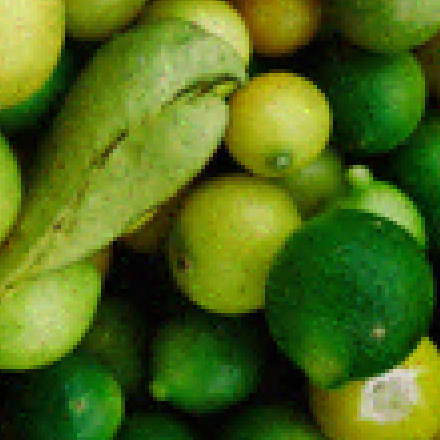} 
                    \hspace{\fsdurthree} &
                    \includegraphics[width= \textwidth]{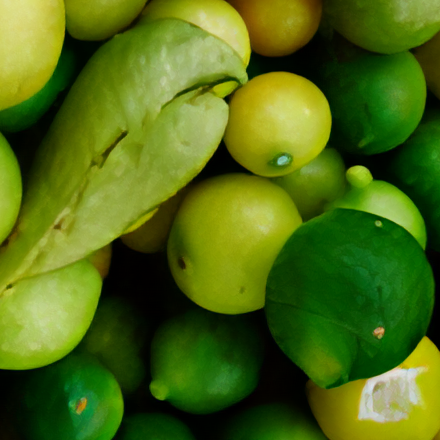} \hspace{\fsdurthree} &
                    \includegraphics[width= \textwidth]{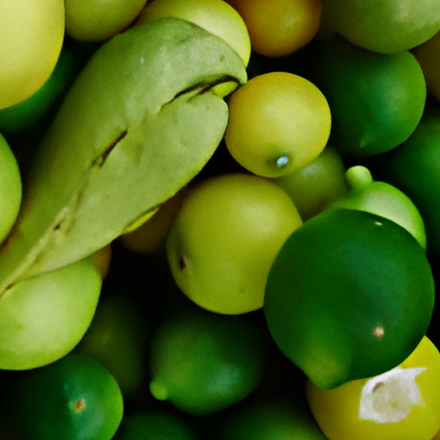} \hspace{\fsdurthree} &
                    \includegraphics[width= \textwidth]{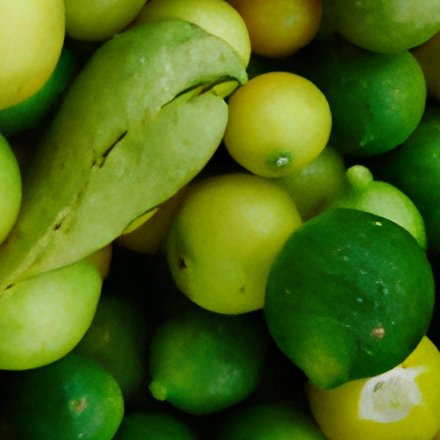} \hspace{\fsdurthree}  
                    \\ 
                    \Huge \scalebox{1.7}{LQ} \hspace{\fsdurthree} &
                    \Huge \scalebox{1.7}{BSRGAN} \hspace{\fsdurthree} &
                    \Huge \scalebox{1.7}{MM-RealSR} \hspace{\fsdurthree} &
                    \makecell{\Huge \scalebox{1.7}{VmambaIR (Ours)}} \hspace{\fsdurthree} 
                \end{tabular}
            \end{adjustbox}

            \\
            \begin{adjustbox}{valign=t}
                \Large
                \begin{tabular}{c}
                    \includegraphics[height=2.15\textwidth]{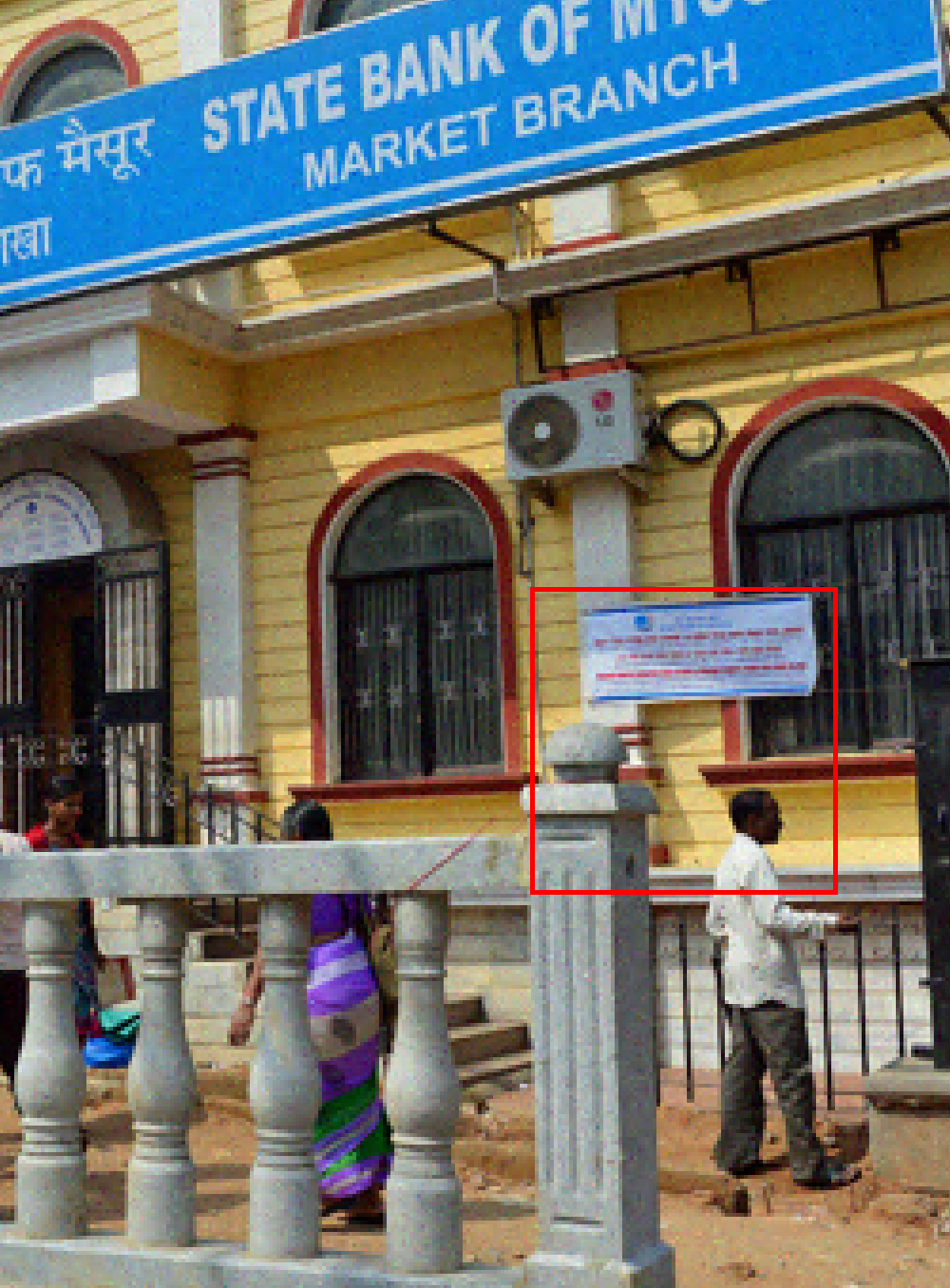} 
                \end{tabular}
                
            \end{adjustbox}
            
            \begin{adjustbox}{valign=t}
                \begin{tabular}{cccc}
                    \includegraphics[width= \textwidth]{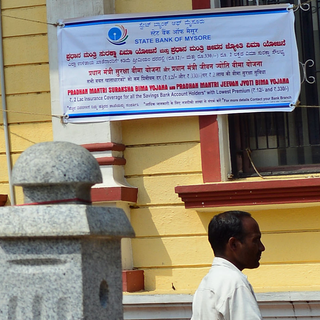} \hspace{\fsdurthree} &
                    \includegraphics[width= \textwidth]{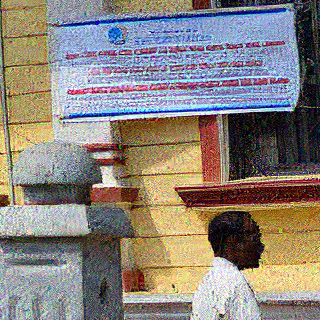} \hspace{\fsdurthree} &
                    \includegraphics[width= \textwidth]{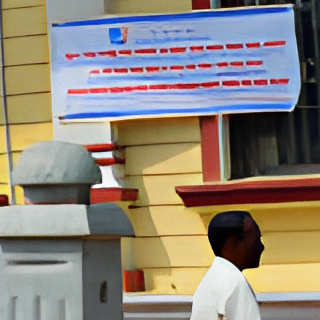} \hspace{\fsdurthree} &
                    \includegraphics[width= \textwidth]{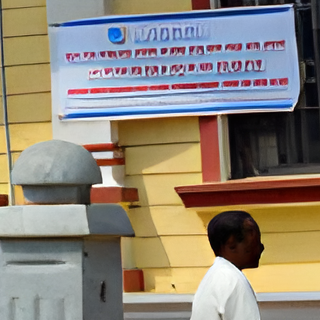} \hspace{\fsdurthree} 
                    \\
                    \Huge \scalebox{1.7}{HQ} \hspace{\fsdurthree} &
                    \Huge \scalebox{1.7}{ESRGAN} \hspace{\fsdurthree} &
                    \makecell{\Huge \scalebox{1.7}{SwinIR}} \hspace{\fsdurthree} &
                    \makecell{\Huge \scalebox{1.7}{Real-ESRGAN}} \hspace{\fsdurthree} 
                    \\
                    \includegraphics[width= \textwidth]{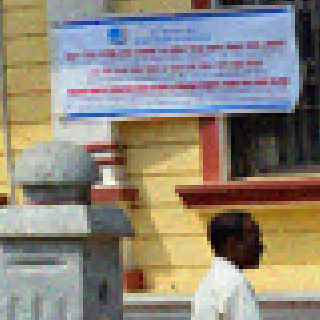} 
                    \hspace{\fsdurthree} &
                    \includegraphics[width= \textwidth]{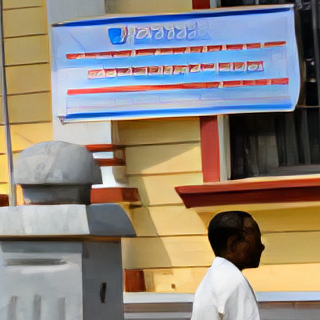} \hspace{\fsdurthree} &
                    \includegraphics[width= \textwidth]{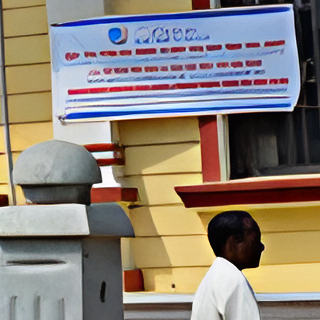} \hspace{\fsdurthree} &
                    \includegraphics[width= \textwidth]{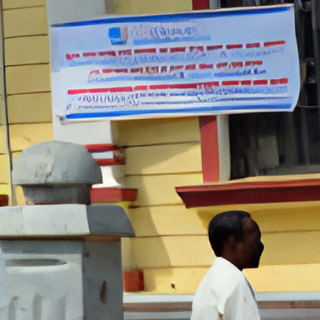} \hspace{\fsdurthree}  
                    \\ 
                    \Huge \scalebox{1.7}{LQ} \hspace{\fsdurthree} &
                    \Huge \scalebox{1.7}{BSRGAN} \hspace{\fsdurthree} &
                    \Huge \scalebox{1.7}{MM-RealSR} \hspace{\fsdurthree} &
                    \makecell{\Huge \scalebox{1.7}{VmambaIR (Ours)}} \hspace{\fsdurthree} 
                \end{tabular}
            \end{adjustbox}
            \\
            \begin{adjustbox}{valign=t}
                \Large
                \begin{tabular}{c}
                    \includegraphics[height=2.15\textwidth]{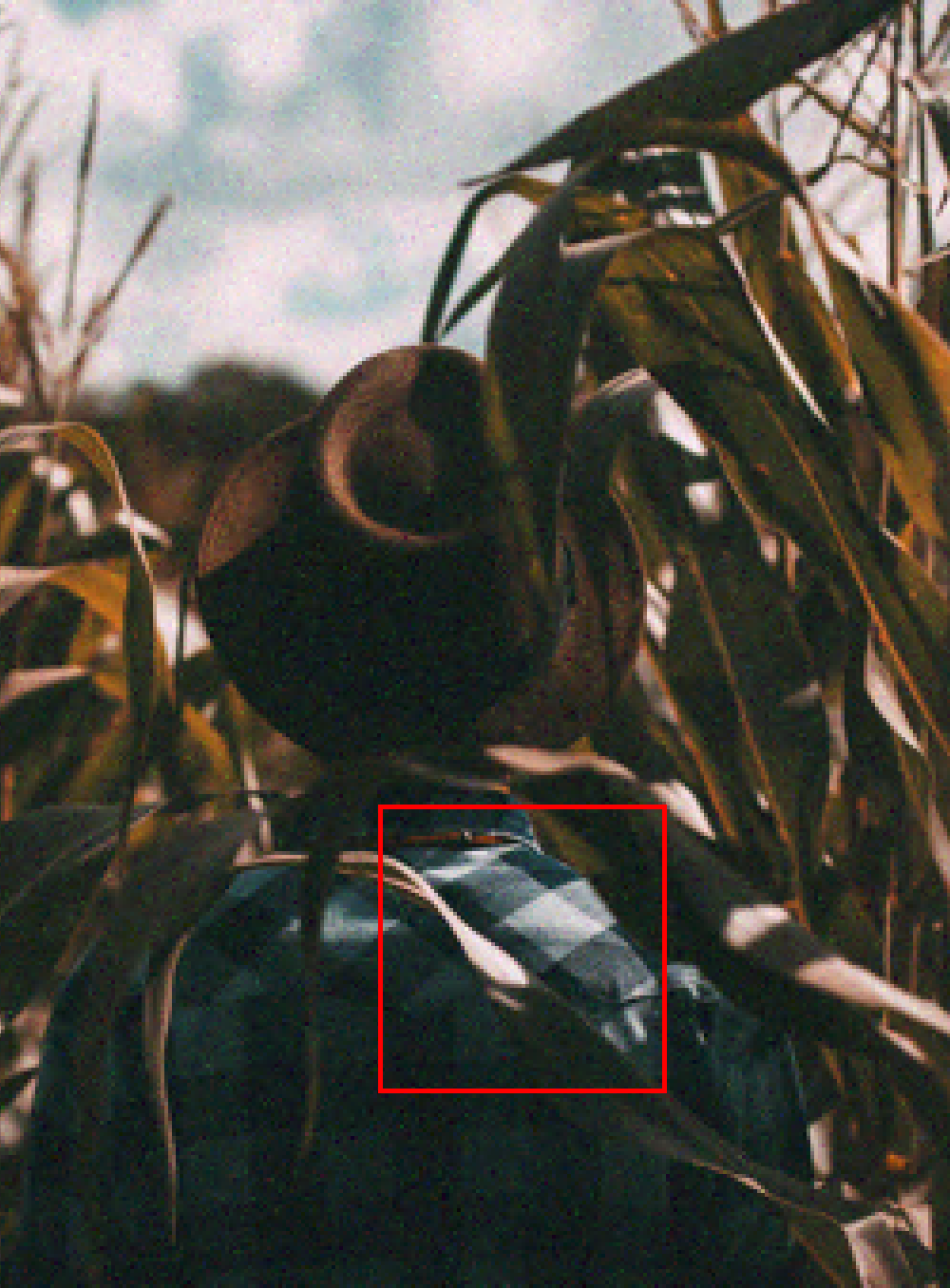} 
                \end{tabular}
                
            \end{adjustbox}
            
            \begin{adjustbox}{valign=t}
                \begin{tabular}{cccc}
                    \includegraphics[width= \textwidth]{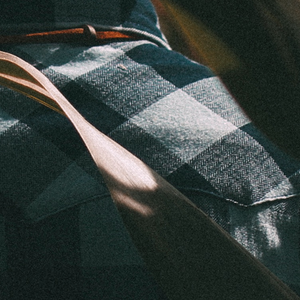} \hspace{\fsdurthree} &
                    \includegraphics[width= \textwidth]{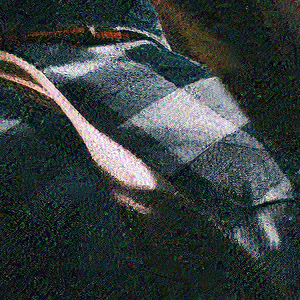} \hspace{\fsdurthree} &
                    \includegraphics[width= \textwidth]{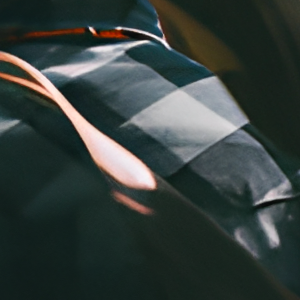} \hspace{\fsdurthree} &
                    \includegraphics[width= \textwidth]{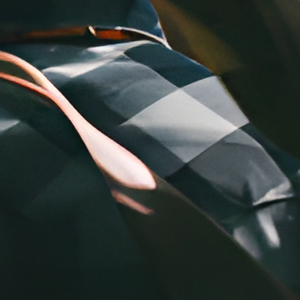} \hspace{\fsdurthree} 
                    \\
                    \Huge \scalebox{1.7}{HQ} \hspace{\fsdurthree} &
                    \Huge \scalebox{1.7}{ESRGAN} \hspace{\fsdurthree} &
                    \makecell{\Huge \scalebox{1.7}{SwinIR}} \hspace{\fsdurthree} &
                    \makecell{\Huge \scalebox{1.7}{Real-ESRGAN}} \hspace{\fsdurthree} 
                    \\
                    \includegraphics[width= \textwidth]{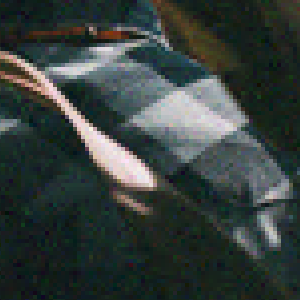} 
                    \hspace{\fsdurthree} &
                    \includegraphics[width= \textwidth]{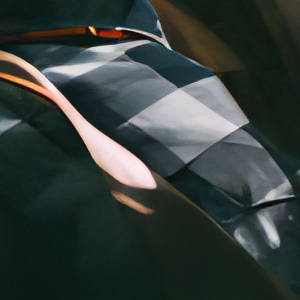} \hspace{\fsdurthree} &
                    \includegraphics[width= \textwidth]{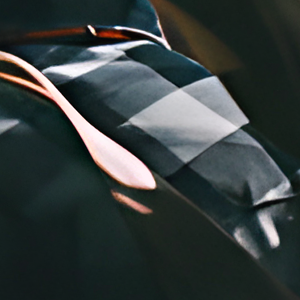} \hspace{\fsdurthree} &
                    \includegraphics[width= \textwidth]{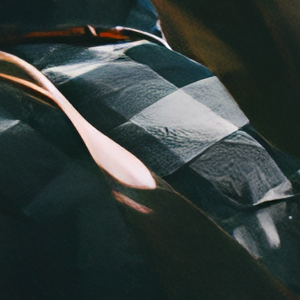} \hspace{\fsdurthree}  
                    \\ 
                    \Huge \scalebox{1.7}{LQ} \hspace{\fsdurthree} &
                    \Huge \scalebox{1.7}{BSRGAN} \hspace{\fsdurthree} &
                    \Huge \scalebox{1.7}{MM-RealSR} \hspace{\fsdurthree} &
                    \makecell{\Huge \scalebox{1.7}{VmambaIR (Ours)}} \hspace{\fsdurthree} 
                \end{tabular}
            \end{adjustbox}
        \\
            \begin{adjustbox}{valign=t}
                \Large
                \begin{tabular}{c}
                    \includegraphics[height=2.15\textwidth]{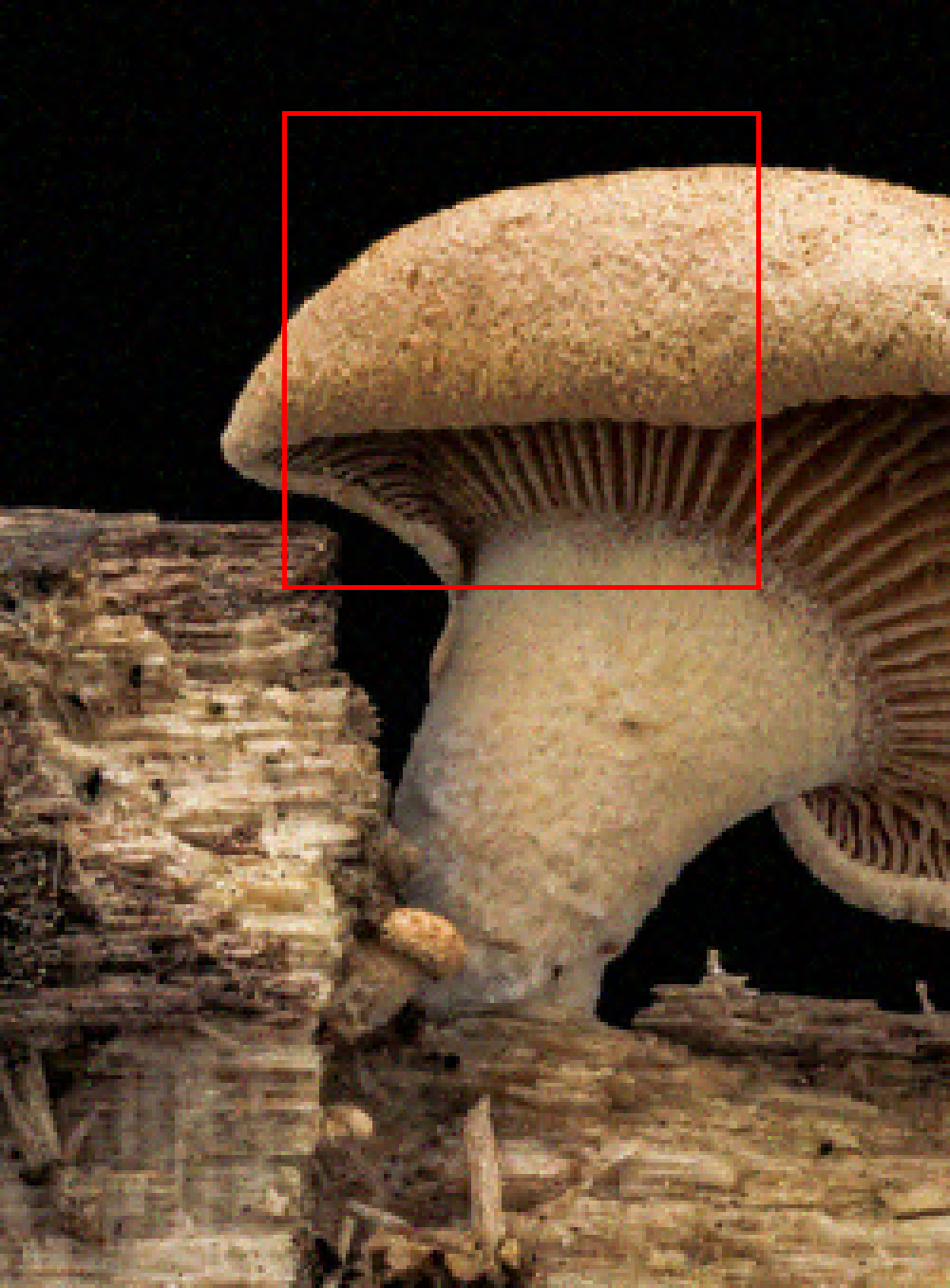} 
                \end{tabular}
                
            \end{adjustbox}
            
            \begin{adjustbox}{valign=t}
                \begin{tabular}{cccc}
                    \includegraphics[width= \textwidth]{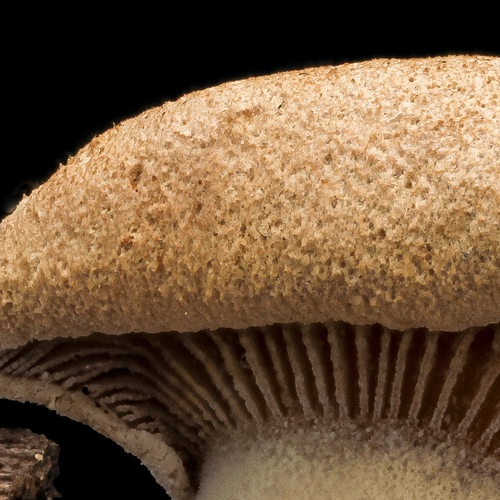} \hspace{\fsdurthree} &
                    \includegraphics[width= \textwidth]{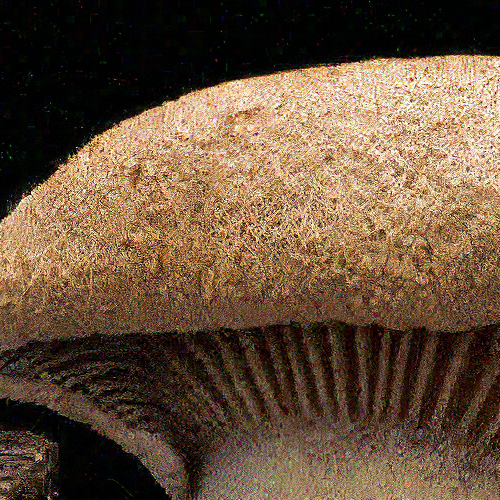} \hspace{\fsdurthree} &
                    \includegraphics[width= \textwidth]{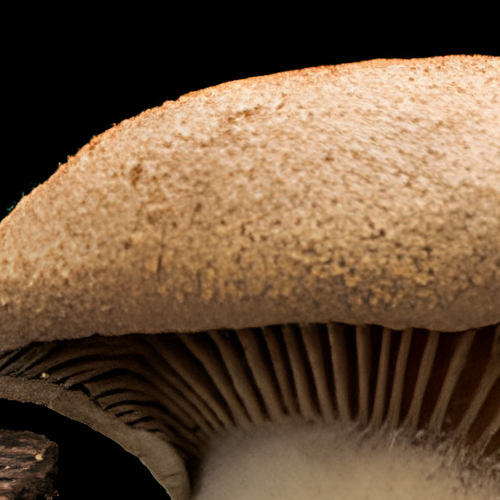} \hspace{\fsdurthree} &
                    \includegraphics[width= \textwidth]{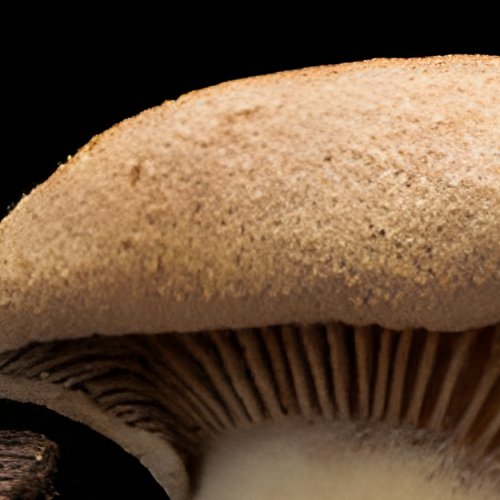} \hspace{\fsdurthree} 
                    \\
                    \Huge \scalebox{1.7}{HQ} \hspace{\fsdurthree} &
                    \Huge \scalebox{1.7}{ESRGAN} \hspace{\fsdurthree} &
                    \makecell{\Huge \scalebox{1.7}{SwinIR}} \hspace{\fsdurthree} &
                    \makecell{\Huge \scalebox{1.7}{Real-ESRGAN}} \hspace{\fsdurthree} 
                    \\
                    \includegraphics[width= \textwidth]{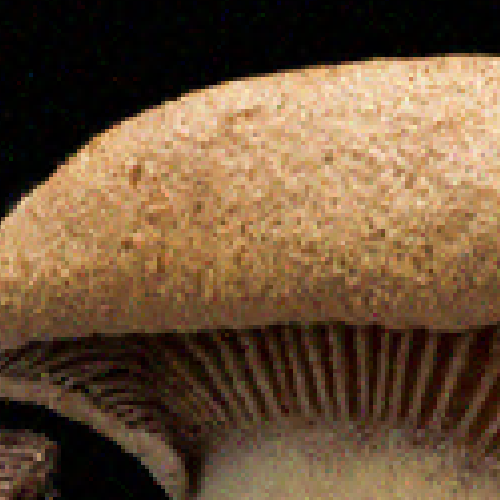} 
                    \hspace{\fsdurthree} &
                    \includegraphics[width= \textwidth]{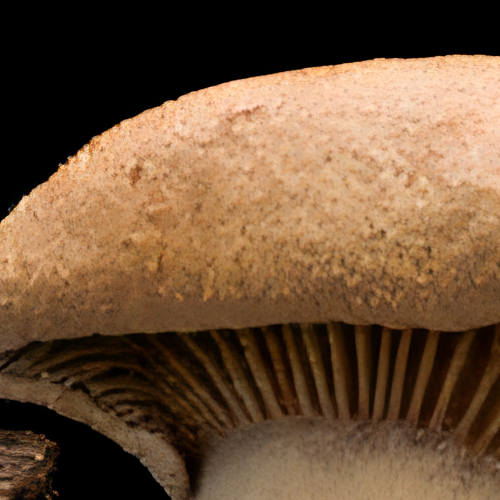} \hspace{\fsdurthree} &
                    \includegraphics[width= \textwidth]{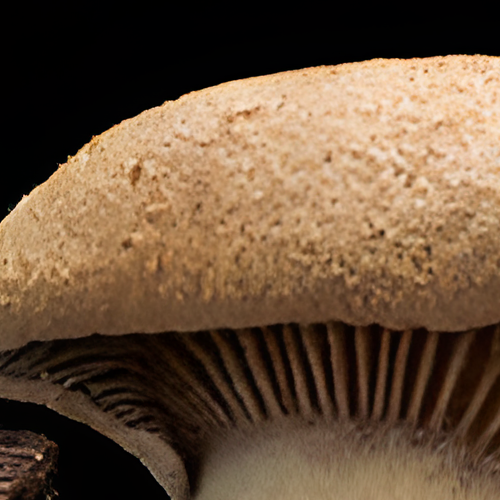} \hspace{\fsdurthree} &
                    \includegraphics[width= \textwidth]{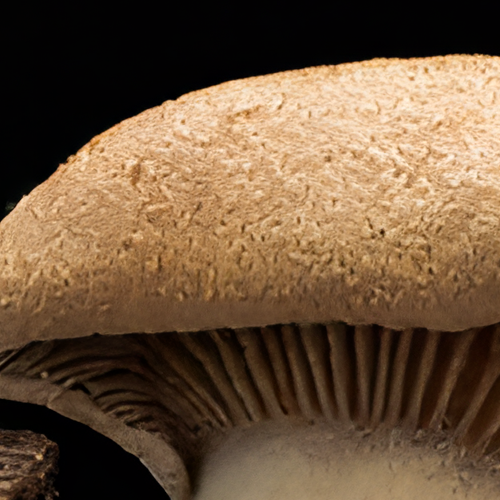} \hspace{\fsdurthree}  
                    \\ 
                    \Huge \scalebox{1.7}{LQ} \hspace{\fsdurthree} &
                    \Huge \scalebox{1.7}{BSRGAN} \hspace{\fsdurthree} &
                    \Huge \scalebox{1.7}{MM-RealSR} \hspace{\fsdurthree} &
                    \makecell{\Huge \scalebox{1.7}{VmambaIR (Ours)}} \hspace{\fsdurthree} 
                \end{tabular}
            \end{adjustbox}
        
        \end{tabular}
    }
    \caption{Visual comparison of \textbf{real-world image super-resolution} methods. Zoom-in for better details.Our VmambaIR achieves superior high-frequency details and fidelity simultaneously.
}
    \label{fig:realsr2}
    \vspace{-3mm}
\end{figure*}

In order to further demonstrate the enhanced fidelity and level of detail exhibited in the images generated by our proposed VmambaIR model for real-world image super-resolution tasks, we present additional visual comparisons in this section. Specifically, we compare the images generated by VmambaIR with those produced by previous SOTA methods such as ESRGAN \cite{zhang2021designing}, BSRGAN \cite{zhang2021designing}, SwinIR \cite{liang2021swinir}, Real-ESRGAN \cite{wang2021real}, and MM-RealSR \cite{mou2022metric}, as illustrated in Figure \ref{fig:realsr2}. Compared to other methods, VmambaIR excels in real-world super-resolution tasks by producing clearer and more accurate high-frequency details while maintaining significantly lower computational overhead.

\subsection{More Visual Comparisons on Single Image Super-Resolution}
\begin{figure*}[h]
    \setlength{\fsdurthree}{0mm}
    \LARGE
    \centering
   \resizebox{0.88\linewidth}{!}{
        \begin{tabular}{c}
            \begin{adjustbox}{valign=t}
                \Large
                \begin{tabular}{c}
                    \includegraphics[height=1.95\textwidth]{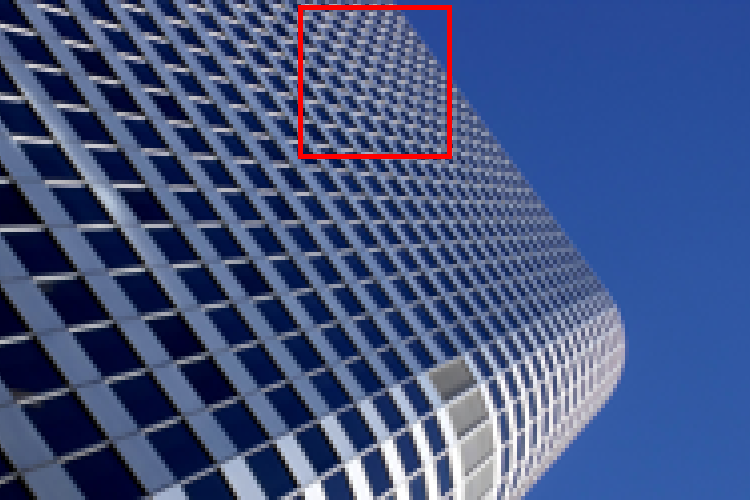} 
                \end{tabular}
                
            \end{adjustbox}
            
            \begin{adjustbox}{valign=t}
                \begin{tabular}{ccc}
                    \includegraphics[width= 0.9\textwidth]{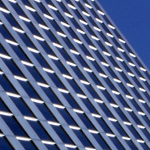} \hspace{\fsdurthree} &
                    \includegraphics[width= 0.9\textwidth]{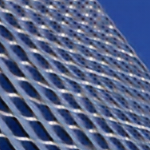} \hspace{\fsdurthree} &
                    \includegraphics[width= 0.9\textwidth]{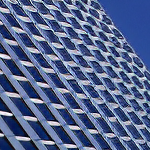} \hspace{\fsdurthree} 
                    \\
                    \Huge \scalebox{1.7}{HQ} \hspace{\fsdurthree} &
                    \makecell{\Huge \scalebox{1.7}{SRGAN}} \hspace{\fsdurthree} &
                    \makecell{\Huge \scalebox{1.7}{ESRGAN}} \hspace{\fsdurthree} 
                    \\
                    \includegraphics[width= 0.9\textwidth]{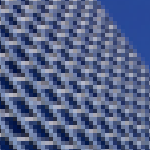} 
                    \hspace{\fsdurthree} &
                    \includegraphics[width= 0.9\textwidth]{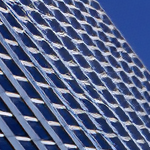} \hspace{\fsdurthree} &
                    \includegraphics[width= 0.9\textwidth]{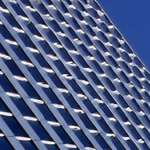} \hspace{\fsdurthree}  
                    \\ 
                    \Huge \scalebox{1.7}{LQ} \hspace{\fsdurthree} &
                    \Huge \scalebox{1.7}{BebyGAN} \hspace{\fsdurthree} &
                    \makecell{\Huge \scalebox{1.7}{VmambaIR (Ours)}} \hspace{\fsdurthree} 
                \end{tabular}
            \end{adjustbox}

            \\
            \begin{adjustbox}{valign=t}
                \Large
                \begin{tabular}{c}
                    \includegraphics[height=1.95\textwidth]{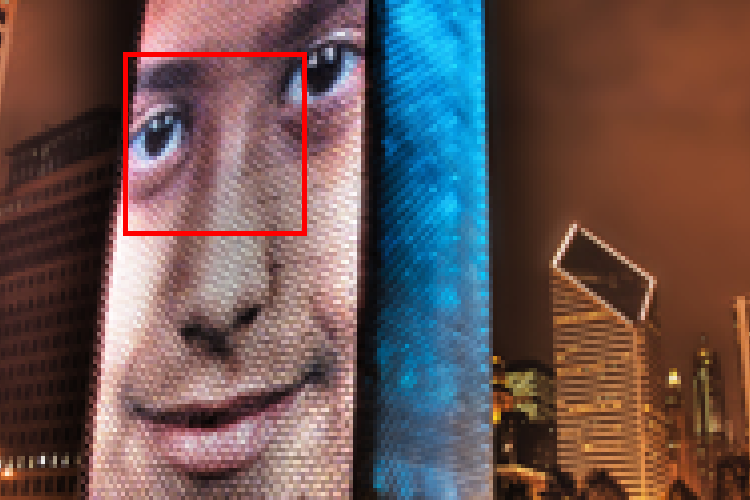} 
                \end{tabular}
                
            \end{adjustbox}
            
            \begin{adjustbox}{valign=t}
                \begin{tabular}{ccc}
                    \includegraphics[width= 0.9\textwidth]{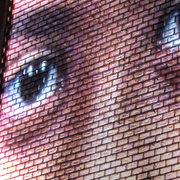} \hspace{\fsdurthree} &
                    \includegraphics[width= 0.9\textwidth]{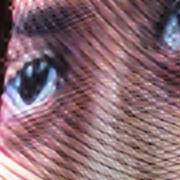} \hspace{\fsdurthree} &
                    \includegraphics[width= 0.9\textwidth]{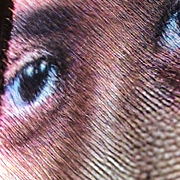} \hspace{\fsdurthree} 
                    \\
                    \Huge \scalebox{1.7}{HQ} \hspace{\fsdurthree} &
                    \makecell{\Huge \scalebox{1.7}{SRGAN}} \hspace{\fsdurthree} &
                    \makecell{\Huge \scalebox{1.7}{ESRGAN}} \hspace{\fsdurthree} 
                    \\
                    \includegraphics[width= 0.9\textwidth]{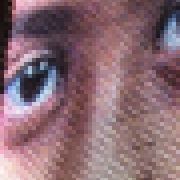} 
                    \hspace{\fsdurthree} &
                    \includegraphics[width= 0.9\textwidth]{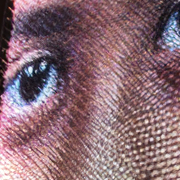} \hspace{\fsdurthree} &
                    \includegraphics[width= 0.9\textwidth]{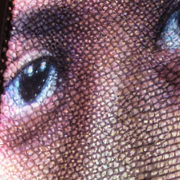} \hspace{\fsdurthree}  
                    \\ 
                    \Huge \scalebox{1.7}{LQ} \hspace{\fsdurthree} &
                    \Huge \scalebox{1.7}{BebyGAN} \hspace{\fsdurthree} &
                    \makecell{\Huge \scalebox{1.7}{VmambaIR (Ours)}} \hspace{\fsdurthree} 
                \end{tabular}
            \end{adjustbox}
            \\
            \begin{adjustbox}{valign=t}
                \Large
                \begin{tabular}{c}
                    \includegraphics[height=1.95\textwidth]{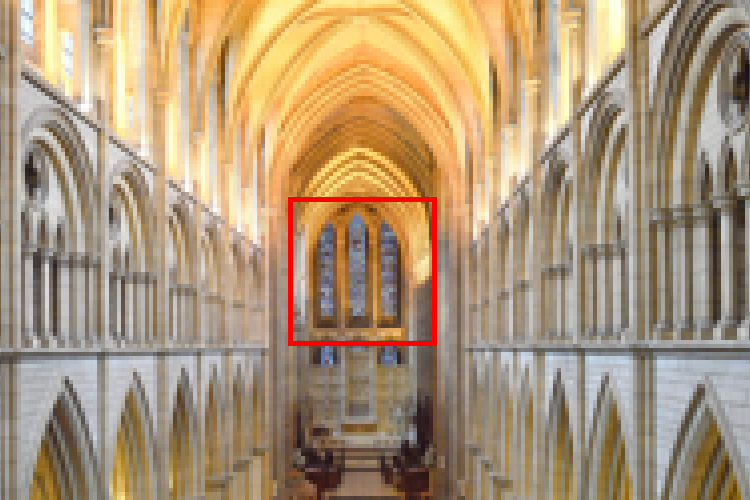} 
                \end{tabular}
                
            \end{adjustbox}
            
            \begin{adjustbox}{valign=t}
                \begin{tabular}{ccc}
                    \includegraphics[width= 0.9\textwidth]{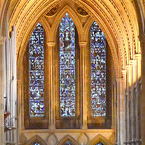} \hspace{\fsdurthree} &
                    \includegraphics[width= 0.9\textwidth]{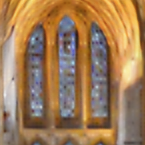} \hspace{\fsdurthree} &
                    \includegraphics[width= 0.9\textwidth]{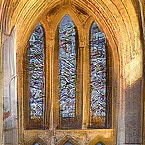} \hspace{\fsdurthree} 
                    \\
                    \Huge \scalebox{1.7}{HQ} \hspace{\fsdurthree} &
                    \makecell{\Huge \scalebox{1.7}{SRGAN}} \hspace{\fsdurthree} &
                    \makecell{\Huge \scalebox{1.7}{ESRGAN}} \hspace{\fsdurthree} 
                    \\
                    \includegraphics[width= 0.9\textwidth]{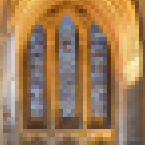} 
                    \hspace{\fsdurthree} &
                    \includegraphics[width= 0.9\textwidth]{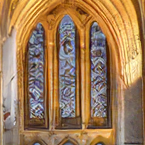} \hspace{\fsdurthree} &
                    \includegraphics[width= 0.9\textwidth]{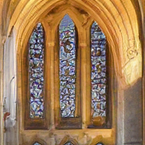} \hspace{\fsdurthree}  
                    \\ 
                    \Huge \scalebox{1.7}{LQ} \hspace{\fsdurthree} &
                    \Huge \scalebox{1.7}{BebyGAN} \hspace{\fsdurthree} &
                    \makecell{\Huge \scalebox{1.7}{VmambaIR (Ours)}} \hspace{\fsdurthree} 
                \end{tabular}
            \end{adjustbox}
        \\
            \begin{adjustbox}{valign=t}
                \Large
                \begin{tabular}{c}
                    \includegraphics[height=1.95\textwidth]{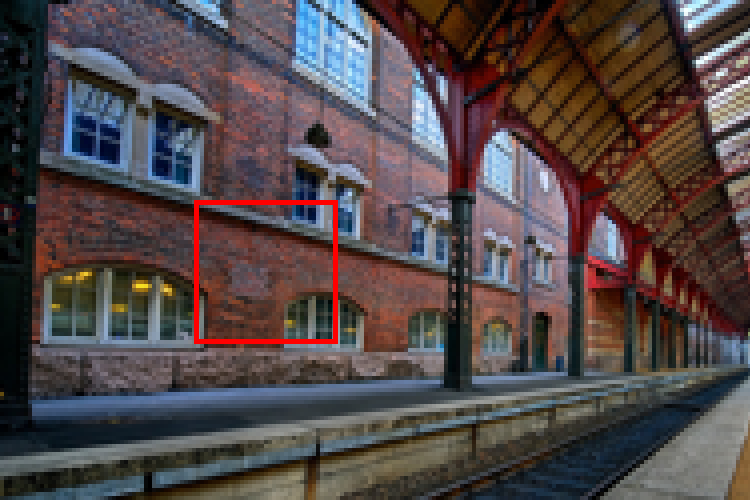} 
                \end{tabular}
                
            \end{adjustbox}
            
            \begin{adjustbox}{valign=t}
                \begin{tabular}{ccc}
                    \includegraphics[width= 0.9\textwidth]{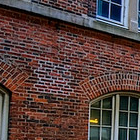} \hspace{\fsdurthree} &
                    \includegraphics[width= 0.9\textwidth]{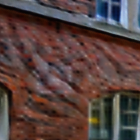} \hspace{\fsdurthree} &
                    \includegraphics[width= 0.9\textwidth]{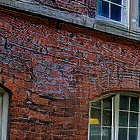} \hspace{\fsdurthree} 
                    \\
                    \Huge \scalebox{1.7}{HQ} \hspace{\fsdurthree} &
                    \makecell{\Huge \scalebox{1.7}{SRGAN}} \hspace{\fsdurthree} &
                    \makecell{\Huge \scalebox{1.7}{ESRGAN}} \hspace{\fsdurthree} 
                    \\
                    \includegraphics[width= 0.9\textwidth]{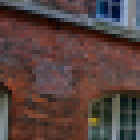} 
                    \hspace{\fsdurthree} &
                    \includegraphics[width= 0.9\textwidth]{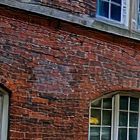} \hspace{\fsdurthree} &
                    \includegraphics[width= 0.9\textwidth]{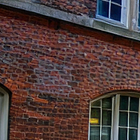} \hspace{\fsdurthree}  
                    \\ 
                    \Huge \scalebox{1.7}{LQ} \hspace{\fsdurthree} &
                    \Huge \scalebox{1.7}{BebyGAN} \hspace{\fsdurthree} &
                    \makecell{\Huge \scalebox{1.7}{VmambaIR (Ours)}} \hspace{\fsdurthree} 
                \end{tabular}
            \end{adjustbox}
            \\
            \begin{adjustbox}{valign=t}
                \Large
                \begin{tabular}{c}
                    \includegraphics[height=1.95\textwidth]{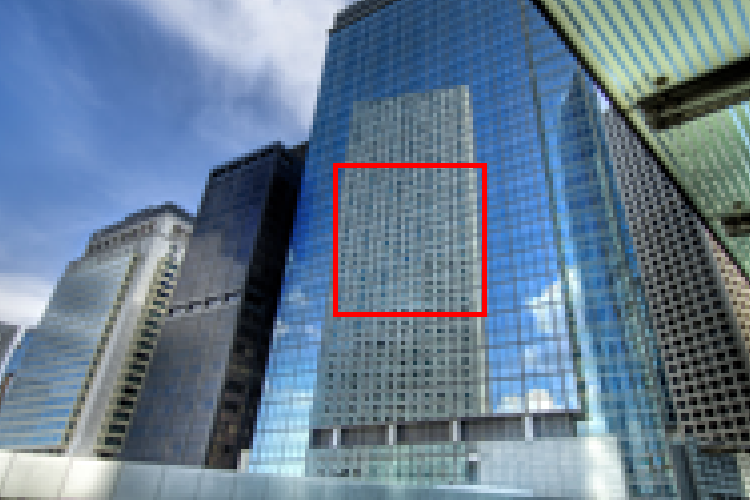} 
                \end{tabular}
                
            \end{adjustbox}
            
            \begin{adjustbox}{valign=t}
                \begin{tabular}{ccc}
                    \includegraphics[width= 0.9\textwidth]{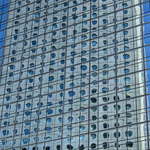} \hspace{\fsdurthree} &
                    \includegraphics[width= 0.9\textwidth]{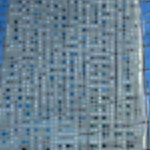} \hspace{\fsdurthree} &
                    \includegraphics[width= 0.9\textwidth]{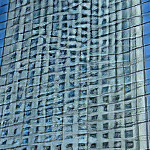} \hspace{\fsdurthree} 
                    \\
                    \Huge \scalebox{1.7}{HQ} \hspace{\fsdurthree} &
                    \makecell{\Huge \scalebox{1.7}{SRGAN}} \hspace{\fsdurthree} &
                    \makecell{\Huge \scalebox{1.7}{ESRGAN}} \hspace{\fsdurthree} 
                    \\
                    \includegraphics[width= 0.9\textwidth]{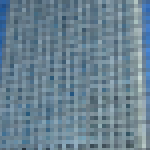} 
                    \hspace{\fsdurthree} &
                    \includegraphics[width= 0.9\textwidth]{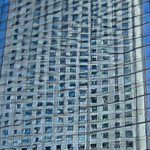} \hspace{\fsdurthree} &
                    \includegraphics[width= 0.9\textwidth]{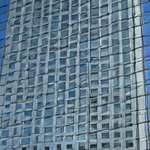} \hspace{\fsdurthree}  
                    \\ 
                    \Huge \scalebox{1.7}{LQ} \hspace{\fsdurthree} &
                    \Huge \scalebox{1.7}{BebyGAN} \hspace{\fsdurthree} &
                    \makecell{\Huge \scalebox{1.7}{VmambaIR (Ours)}} \hspace{\fsdurthree} 
                \end{tabular}
            \end{adjustbox}
        
        \end{tabular}
    }
    \caption{Visual comparison of \textbf{single image super-resolution} methods. Zoom-in for better details.
}
    \label{fig:sisr2}
    \vspace{-3mm}
\end{figure*}

In order to further demonstrate the enhanced fidelity and level of detail exhibited in the images generated by our proposed VmambaIR model for single image super-resolution tasks, we present additional visual comparisons in this section. Specifically, we compare the images generated by VmambaIR with those produced by previous SOTA methods such as SRGAN \cite{ledig2017photo},  ESRGAN \cite{zhang2021designing}, and BebyGAN \cite{li2022best}, as illustrated in Figure \ref{fig:sisr2}. In the single image super-resolution task, our VmambaIR exhibits remarkable proficiency in restoring fine details across a broader range of test images. This serves as a testament to the robust high-frequency modeling capability of mamba \cite{gu2023mamba} in image restoration. 

\subsection{More Visual Comparisons on Image Deraining}
\begin{figure*}[htbp]
    \setlength{\fsdurthree}{0mm}
    \Huge
    \centering
   \resizebox{1\linewidth}{!}{
            \begin{adjustbox}{valign=t}
                \begin{tabular}{cccccc}

                    \includegraphics[width= \textwidth]{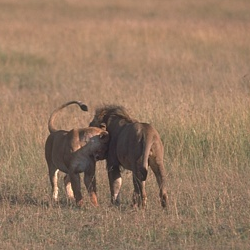} \hspace{\fsdurthree} &
                    \includegraphics[width=  \textwidth]{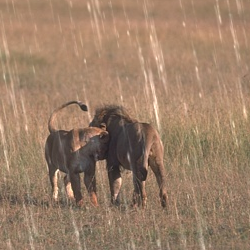} \hspace{\fsdurthree} &
                    \includegraphics[width=  \textwidth]{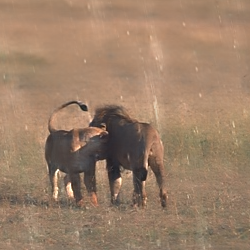} \hspace{\fsdurthree} &
                    \includegraphics[width=  \textwidth]{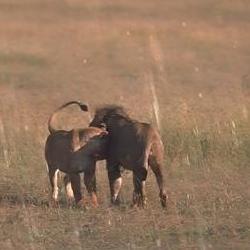} \hspace{\fsdurthree} &
                    \includegraphics[width= \textwidth]{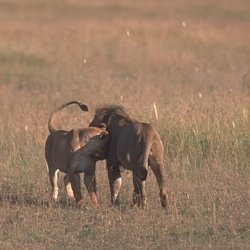} \hspace{\fsdurthree} &
                    \includegraphics[width=  \textwidth]{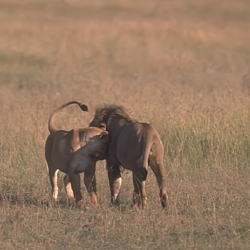} \hspace{\fsdurthree} 
                    \\
                    \includegraphics[width= \textwidth]{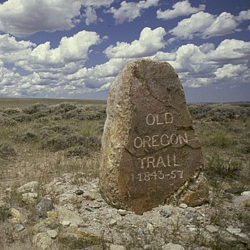} \hspace{\fsdurthree} &
                    \includegraphics[width=  \textwidth]{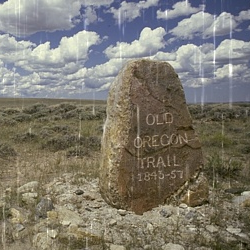} \hspace{\fsdurthree} &
                    \includegraphics[width=  \textwidth]{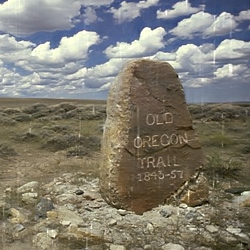} \hspace{\fsdurthree} &
                    \includegraphics[width=  \textwidth]{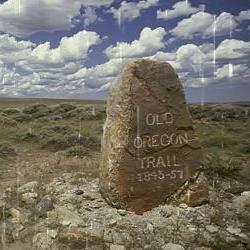} \hspace{\fsdurthree} &
                    \includegraphics[width= \textwidth]{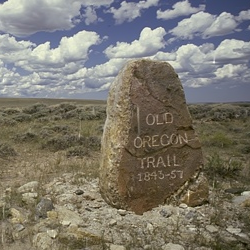} \hspace{\fsdurthree} &
                    \includegraphics[width=  \textwidth]{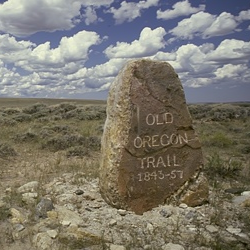} \hspace{\fsdurthree} 
                    \\
                    \includegraphics[width= \textwidth]{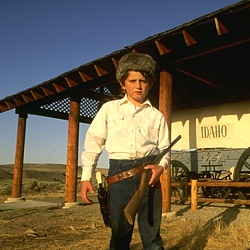} \hspace{\fsdurthree} &
                    \includegraphics[width=  \textwidth]{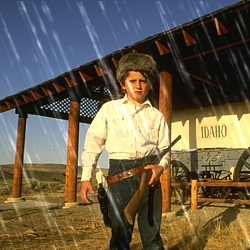} \hspace{\fsdurthree} &
                    \includegraphics[width=  \textwidth]{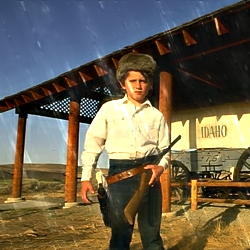} \hspace{\fsdurthree} &
                    \includegraphics[width=  \textwidth]{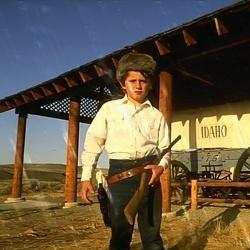} \hspace{\fsdurthree} &
                    \includegraphics[width= \textwidth]{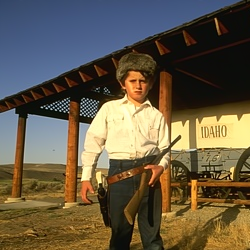} \hspace{\fsdurthree} &
                    \includegraphics[width=  \textwidth]{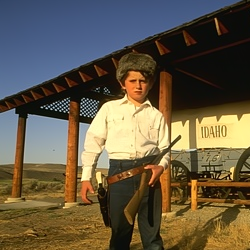} \hspace{\fsdurthree}  
                    \\
                    \includegraphics[width= \textwidth]{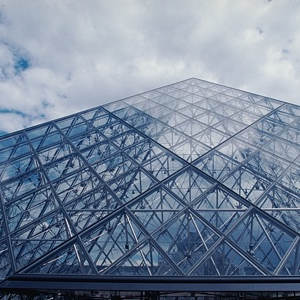} \hspace{\fsdurthree} &
                    \includegraphics[width=  \textwidth]{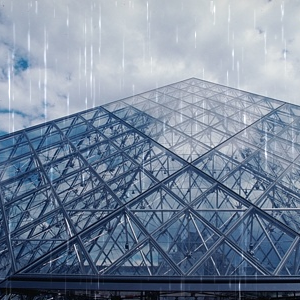} \hspace{\fsdurthree} &
                    \includegraphics[width=  \textwidth]{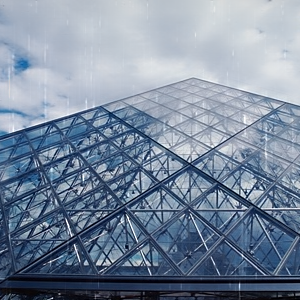} \hspace{\fsdurthree} &
                    \includegraphics[width=  \textwidth]{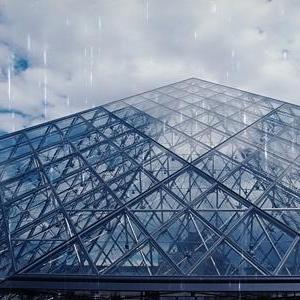} \hspace{\fsdurthree} &
                    \includegraphics[width= \textwidth]{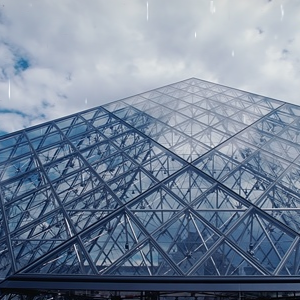} \hspace{\fsdurthree} &
                    \includegraphics[width=  \textwidth]{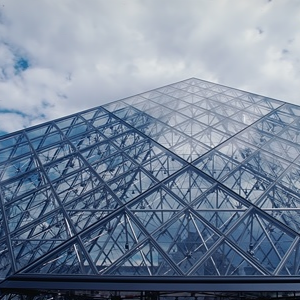} \hspace{\fsdurthree} 
                    \\
                    \includegraphics[width= \textwidth]{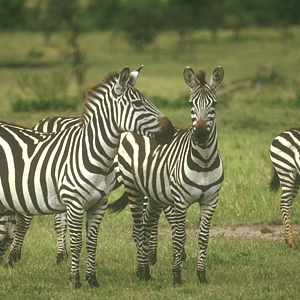} \hspace{\fsdurthree} &
                    \includegraphics[width=  \textwidth]{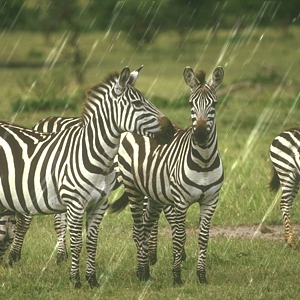} \hspace{\fsdurthree} &
                    \includegraphics[width=  \textwidth]{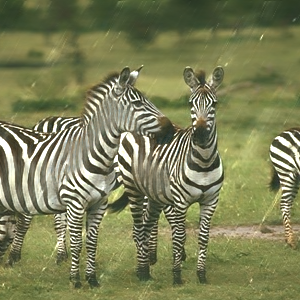} \hspace{\fsdurthree} &
                    \includegraphics[width=  \textwidth]{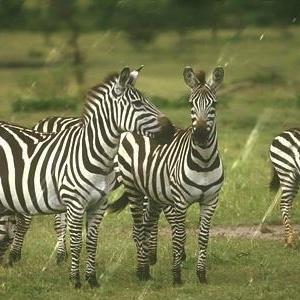} \hspace{\fsdurthree} &
                    \includegraphics[width= \textwidth]{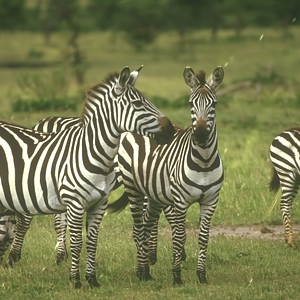} \hspace{\fsdurthree} &
                    \includegraphics[width=  \textwidth]{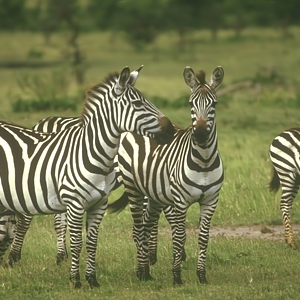} \hspace{\fsdurthree} 
                    \\
                    \includegraphics[width= \textwidth]{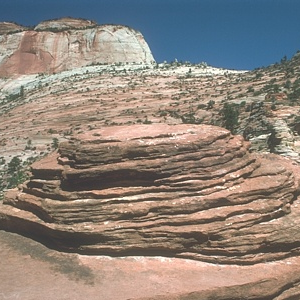} \hspace{\fsdurthree} &
                    \includegraphics[width=  \textwidth]{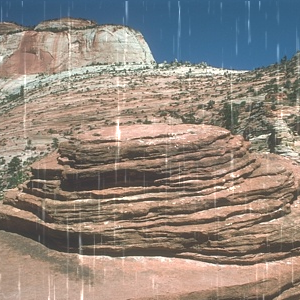} \hspace{\fsdurthree} &
                    \includegraphics[width=  \textwidth]{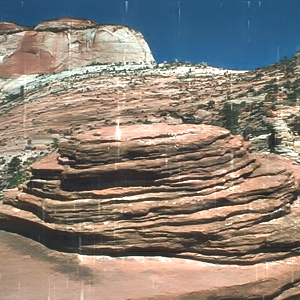} \hspace{\fsdurthree} &
                    \includegraphics[width=  \textwidth]{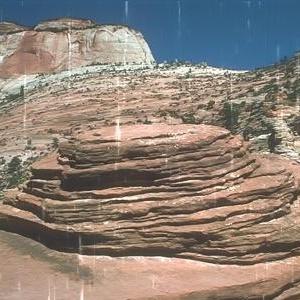} \hspace{\fsdurthree} &
                    \includegraphics[width= \textwidth]{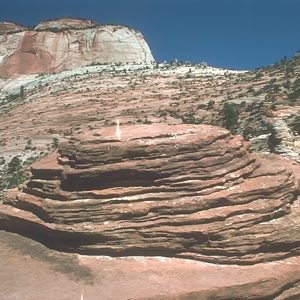} \hspace{\fsdurthree} &
                    \includegraphics[width=  \textwidth]{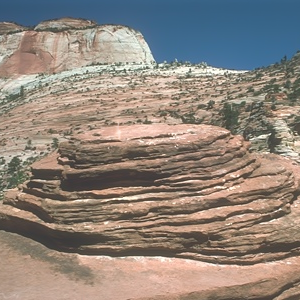} \hspace{\fsdurthree} 
                    \\
                    \includegraphics[width= \textwidth]{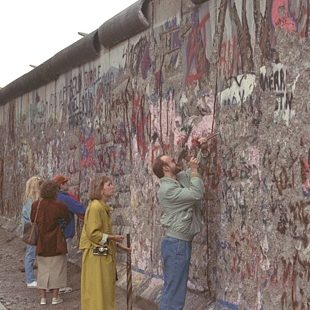} \hspace{\fsdurthree} &
                    \includegraphics[width=  \textwidth]{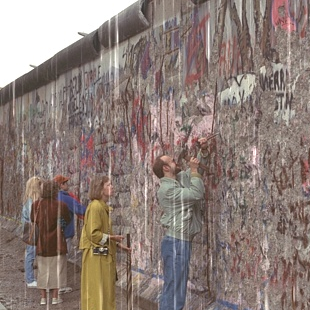} \hspace{\fsdurthree} &
                    \includegraphics[width=  \textwidth]{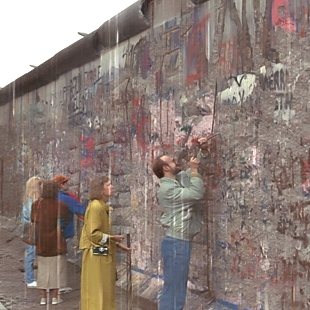} \hspace{\fsdurthree} &
                    \includegraphics[width=  \textwidth]{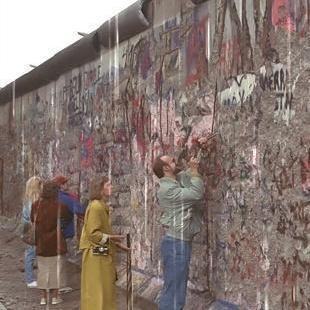} \hspace{\fsdurthree} &
                    \includegraphics[width= \textwidth]{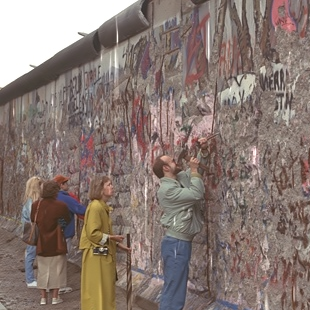} \hspace{\fsdurthree} &
                    \includegraphics[width=  \textwidth]{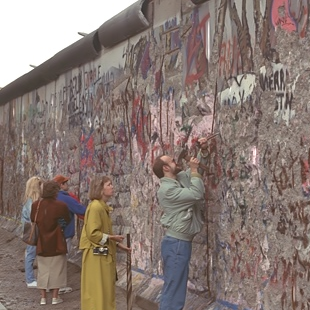} \hspace{\fsdurthree} 
                    \\
                    \scalebox{1.7}{HQ} \hspace{\fsdurthree} &
                    \scalebox{1.7}{LQ} \hspace{\fsdurthree} &
                    \scalebox{1.7}{\makecell{DerainNet~\cite{fu2017clearing}}} \hspace{\fsdurthree} &
                    \scalebox{1.7}{RESCAN~\cite{li2018recurrent}} \hspace{\fsdurthree} &
                    \scalebox{1.7}{Restormer~\cite{zamir2022restormer}} \hspace{\fsdurthree} &
                    \makecell{\scalebox{1.7}{VmambaIR (Ours)}} \hspace{\fsdurthree} 
                \end{tabular}
            \end{adjustbox}

    }
    \caption{Visual comparison of \textbf{image deraining} methods. Zoom-in for better details.}
    \vspace{-3mm}
    \label{fig:derain2}
\end{figure*}

In order to further demonstrate the enhanced fidelity and level of detail exhibited in the images generated by our proposed VmambaIR model for image deraining tasks, we present additional visual comparisons in this section. Specifically, we compare the images generated by VmambaIR with those produced by previous SOTA methods such as Restormer \cite{zamir2022restormer}, RESCAN \cite{li2018recurrent}, DerainNet \cite{fu2017clearing}, as illustrated in Figure \ref{fig:derain2}.
In the image deraining task, our VmambaIR consistently achieves near-perfect restoration results across a diverse set of images. When compared to previous state-of-the-art methods \cite{zamir2022restormer, li2018recurrent}, our VmambaIR showcases distinct visual advantages and delivers superior performance, solidifying its position as a leading approach in the field.

\end{document}